\documentclass[11pt]{article}

\usepackage[final]{acl}

\usepackage{times}
\usepackage{latexsym}

\usepackage[T1]{fontenc}

\usepackage[utf8]{inputenc}

\usepackage{microtype}

\usepackage{inconsolata}

\usepackage{graphicx}

\usepackage{color}
\usepackage{multirow}
\usepackage{tcolorbox}
\tcbuselibrary{skins, breakable}
\usepackage{newfloat}
\usepackage{listings}
\usepackage{algorithm}
\usepackage{algorithmic}
\usepackage{booktabs}
\usepackage{amsmath}
\usepackage{amssymb}
\usepackage{makecell}
\usepackage{pifont}  
\usepackage{xcolor}
\usepackage{soul}
\usepackage{enumitem}
\usepackage{hyperref}
\usepackage[table]{xcolor} 
\usepackage[dvipsnames]{xcolor}
\usepackage{pgf}

\setlist[itemize]{leftmargin=*}
\setitemize[1]{itemsep=0pt,partopsep=0pt,parsep=\parskip,topsep=0pt}

\newcommand{\boldpara}[1]{%
  \par\vspace{1ex}\noindent\textbf{#1}\space
}

\DeclareRobustCommand{\logopsyche}{%
    \raisebox{-0.15\height}{%
        \includegraphics[height=1em]{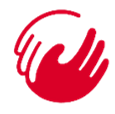}
    }
    \hspace{-0.6em}
}

\title{\logopsyche Psyche-R1: Towards Reliable Psychological LLMs through Unified Empathy, Expertise, and Reasoning}

\author{
 \textbf{Chongyuan Dai\textsuperscript{1}},
 \textbf{Jinpeng Hu\textsuperscript{1$\dagger$}},
 \textbf{Hongchang Shi\textsuperscript{1}},
 \textbf{Zhuo Li\textsuperscript{2}},
 \textbf{Dan Guo\textsuperscript{1}},
\\
 \textbf{Xun Yang\textsuperscript{3}},
 \textbf{Meng Wang\textsuperscript{1}}
\\
\\
 \textsuperscript{1}Hefei University of Technology,
 \textsuperscript{2}The Chinese University of Hong Kong, Shenzhen,
\\
 \textsuperscript{3}University of Science and Technology of China,
\\
\texttt{2023217261@mail.hfut.edu.cn, jinpenghu@hfut.edu.cn}
}

\begin{document}
\maketitle

\begingroup
    \renewcommand\thefootnote{}
    \footnotetext{$^\dagger$Corresponding author}
\endgroup

\begin{abstract}
Amidst a shortage of qualified mental health professionals, the integration of large language models (LLMs) into psychological applications offers a promising way to alleviate the growing burden of mental health disorders. Recent reasoning-augmented LLMs have achieved remarkable performance in mathematics and programming, while research in the psychological domain has predominantly emphasized emotional support and empathetic dialogue, with limited attention to reasoning mechanisms that are beneficial to generating accurate responses. Therefore, in this paper, we propose \logopsyche\textit{Psyche-R1}, the first Chinese psychological LLM that jointly integrates empathy, psychological expertise, and reasoning, built upon a novel data curation pipeline. Specifically, we design a comprehensive data synthesis pipeline that produces over 75k high-quality psychological questions paired with detailed rationales, generated through an iterative prompt-rationale optimization procedure, along with 73k empathetic dialogues. Subsequently, we employ a hybrid training strategy wherein challenging samples are identified through a multi-LLM cross-selection strategy for group relative policy optimization (GRPO) to improve reasoning ability, while the remaining data are used for supervised fine-tuning (SFT) to enhance empathetic response generation and psychological domain knowledge. Extensive experiment results demonstrate the effectiveness of \textit{Psyche-R1} across several psychological benchmarks, where our 7B \textit{Psyche-R1} achieves comparable results to 671B \texttt{DeepSeek-R1}\footnote{\url{https://github.com/MindIntLab-HFUT/Psyche-R1}}.

\end{abstract}

\section{Introduction}
The shortage of qualified mental health professionals has spurred increasing interest in applying artificial intelligence within the psychological domain to support mental health assistance \cite{wolohan2018detecting, al2019depression, tanana2021you,10960233,xu-etal-2025-multiagentesc,10.1145/3774904.3792594}.
Recently, large language models (LLMs) have demonstrated impressive capabilities across a wide range of domains owing to their exceptional text understanding capabilities \cite{zhang-etal-2023-huatuogpt, naveed2023comprehensive, shi2026codeocreffectivenessvisionlanguage,zhang2026cogevolutionhumanlikegenerativeeducational}.
Therefore, many LLM-based studies have been proposed to advance the mental health services \cite{cho2023evaluating, ye-etal-2025-sweetiechat}.

Prior research has established the critical importance of empathy optimization in psychological counseling \cite{qiu-etal-2024-smile, sorin2024large, zhang-etal-2024-cpsycoun}. 
\begin{figure}[t]
\centering
\includegraphics[width=\columnwidth]{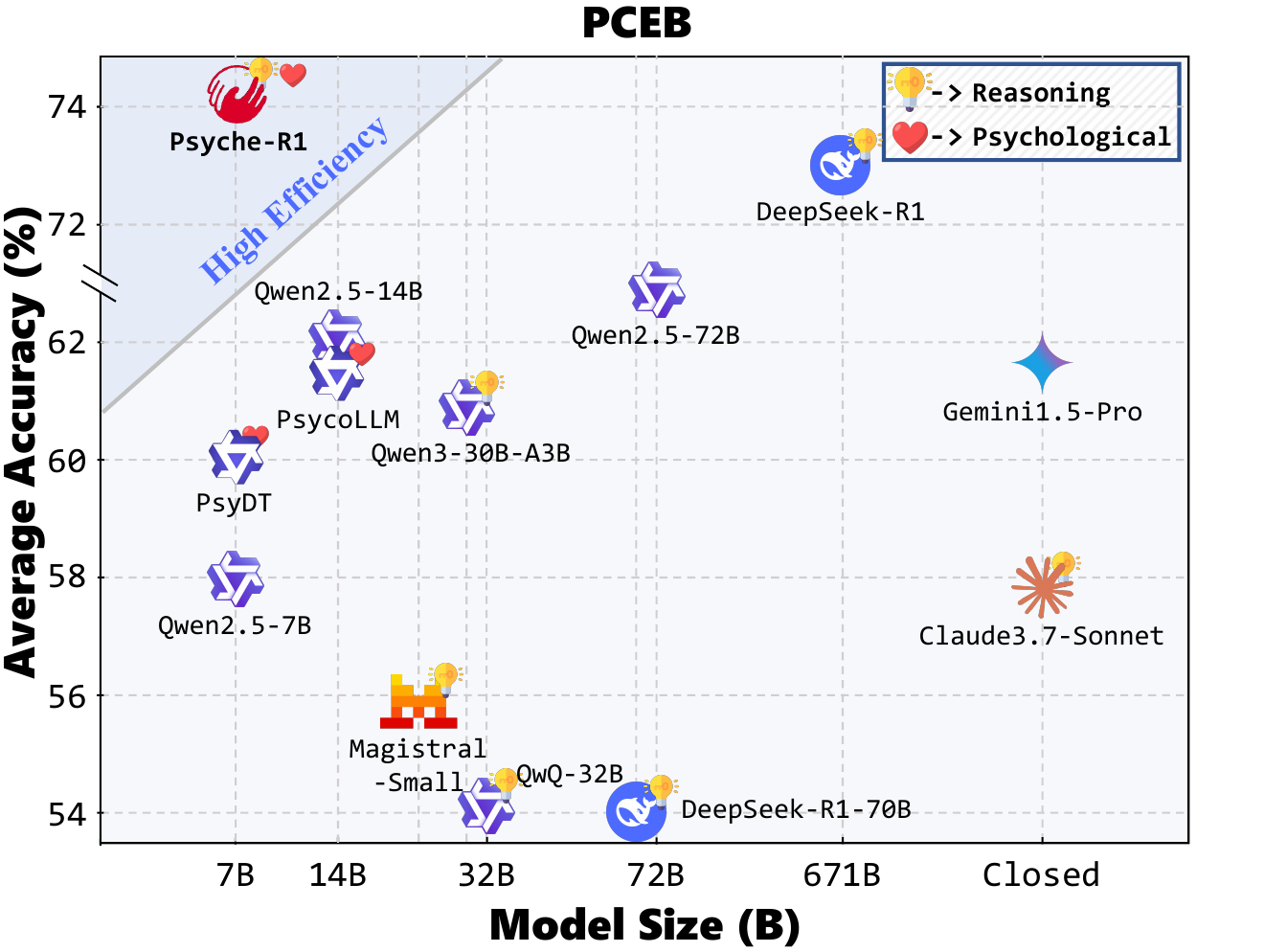}
\caption{Comparison of different LLMs on the PCEB, plotted by average standard accuracy versus model size.}
\label{fig_intro}
\vspace{-1em}
\end{figure}
For example, SoulChat \cite{chen-etal-2023-soulchat} enhances empathetic responding by fine-tuning a model on a large-scale, multi-turn empathetic dialogue dataset. 
Similarly, AUGESC \cite{zheng-etal-2023-augesc} improves emotional sensitivity in dialogue systems by incorporating an emotion-aware attention mechanism.
However, these approaches often lack the expertise foundation required for psychology, which is important for accurate psychological understanding.
Some studies have attempted to address this limitation through integration of psychological knowledge \cite{chen-etal-2023-empowering, xiao-etal-2024-healme, wu2025psychological}. 
\begin{figure*}[t]
\centering
\includegraphics[width=0.97\textwidth]{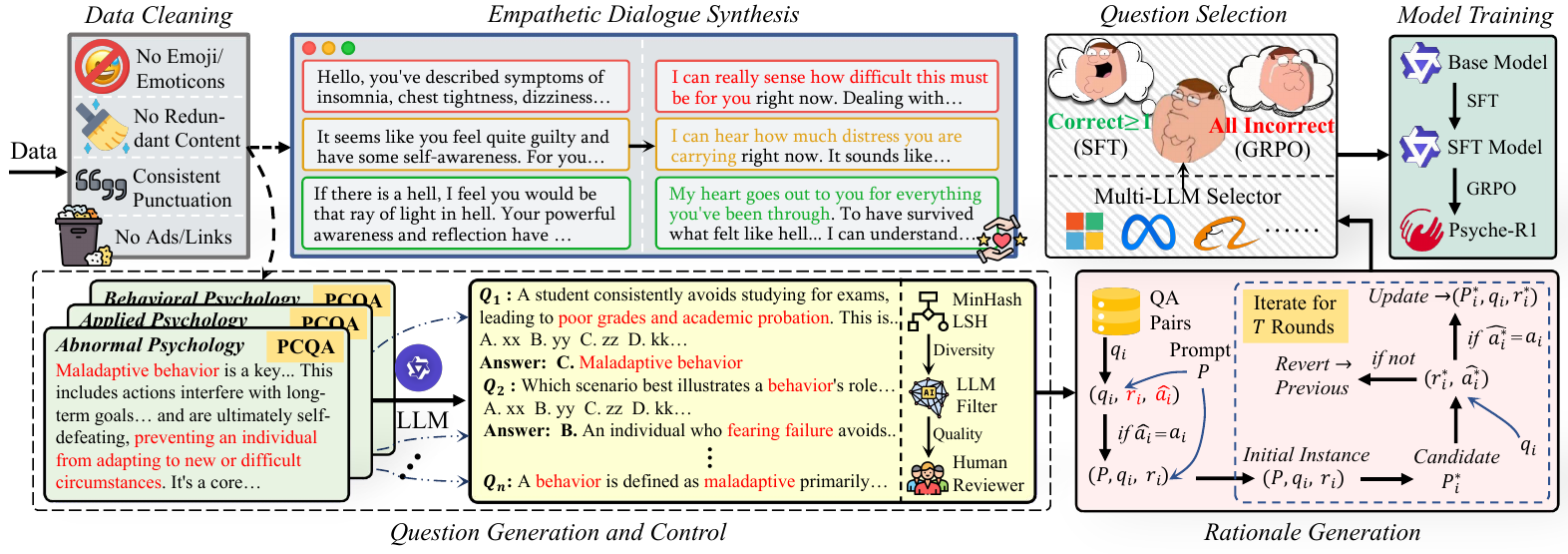} 
\caption{
Overview of our proposed pipeline for constructing the dataset and \logopsyche\textit{Psyche-R1}. Our pipeline involves generating psychological reasoning questions paired with detailed rationales, along with empathetic dialogues.
}
\label{fig2}
\vspace{-1em}
\end{figure*}
For example, \texttt{PsycoLLM} \cite{hu2024psycollm} integrates psychological knowledge by training its model on knowledge-based question-answer (QA) pairs, while CPsyExam \cite{zhao-etal-2025-cpsyexam} leverages examination questions covering theoretical knowledge from different psychology-related subjects to further improve model performance.
Although existing studies have achieved considerable success, they remain limited in their capacity for complex reasoning \cite{hu2025beyond}.
In fact, reasoning-augmented LLMs trained through reinforcement learning (RL) have demonstrated superior performance across various domains, particularly in mathematics, code generation, and medical domain \cite{chen2024huatuogpt, guo2025deepseek, liu2026attention}.
However, as shown in Figure \ref{fig_intro}, these reasoning-augmented LLMs exhibit limited performance in the psychological domain, since they focus on logic reasoning, while neglecting the unification of empathy and expertise beyond general common sense.
In fact, within the psychological domain, reasoning plays a critical role, as it contributes not only to generating more accurate and reliable responses but also to supporting deeper empathetic engagement and more coherent integration of psychological knowledge.

Therefore, in this paper, we introduce \logopsyche\textit{Psyche-R1} that integrates empathy, domain-specific expertise, and reasoning capabilities.
To construct a high-quality training corpus, we design a comprehensive data synthesis pipeline that integrates modified empathetic dialogues derived from authentic sources, which capture diverse and empathetic expressions, with knowledge-centric question-answer pairs that encapsulate psychological expertise.
Specifically, we apply chain-of-thought (CoT) prompting to generate an initial detailed rationale for each question, followed by an iterative prompt-rationale optimization process, aiming to enhance both the coherence of the reasoning and its alignment with the corresponding questions.
In parallel, we synthesize 73k empathetic dialogues drawn from diverse social media sources to strengthen affective understanding and emotional support.
To enhance reasoning, we adopt a multi-LLM cross-selection strategy to categorize questions into challenging and non-challenging subsets based on their inferred complexity. 
The non-challenging subset is used for supervised fine-tuning (SFT) to enhance empathetic response generation and domain knowledge, while the challenging subset is utilized for training with group relative policy optimization (GRPO) to improve the model's reasoning capabilities, with both jointly contributing to the development of the \textit{Psyche-R1}.
Experimental results on a range of psychological benchmarks, including knowledge assessment, case-based analysis, and empathy evaluation, demonstrate the effectiveness of our model, where 7B \textit{Psyche-R1} significantly outperforms models of similar scale and achieves competitive performance relative to substantially larger models such as \texttt{DeepSeek-R1}.

\section{\texorpdfstring{\logopsyche}{Psyche-R1} \textit{Psyche-R1}}

In this section, we give details of data curation and two-stage training paradigm, including data collection (\S\ref{sec:data_collection}), psychological reasoning data synthesis (\S\ref{sec:psychological_data_synthesis}), and empathetic dialogue synthesis (\S\ref{sec:empathetic_dialogue_synthesis}).

\subsection{Data Collection}
\label{sec:data_collection}
\boldpara{Data Resource.} 
To construct a comprehensive and diverse dataset, we curate a wide range of resources:
\begin{itemize}[leftmargin=*]
\item \textbf{Type} \uppercase\expandafter{\romannumeral1}: 
Classic psychology textbooks and curricular materials from psychology programs, all collected from publicly accessible repositories, covering more than 19 subfields and concentrated and systematically organized content, including cognitive psychology, social psychology, etc.
\item \textbf{Type} \uppercase\expandafter{\romannumeral2}: Psychological question banks collected from publicly available Chinese educational platforms via web crawlers and manually curation, encompassing theoretical principles and conceptual knowledge across psychology. 
\item \textbf{Type} \uppercase\expandafter{\romannumeral3}: Synthetic data distilled from \texttt{Qwen2.5-72B-Instruct} \cite{qwen2.5} to supplement underrepresented subfields (e.g., sports psychology) and enhance dataset coverage.
\item \textbf{Type} \uppercase\expandafter{\romannumeral4}: 
Dialogic interactions harvested from established mental health support communities (e.g., \texttt{Yixinli}, \texttt{Jiandanxinli} and \texttt{Zhihu}\footnote{\href{https://www.xinli001.com/}{Yixinli}, \href{https://www.jiandanxinli.com/}{Jiandanxinli}, and \href{https://www.zhihu.com/}{Zhihu}}) dedicated to delivering mental health services.
\end{itemize}
The \textbf{Type} \uppercase\expandafter{\romannumeral1}, \textbf{Type} \uppercase\expandafter{\romannumeral2} and \textbf{Type} \uppercase\expandafter{\romannumeral3} are used to construct the psychological reasoning question-answer (PCQA) dataset, designed to enhance domain-specific knowledge acquisition and reasoning capabilities.
The \textbf{Type} \uppercase\expandafter{\romannumeral4} resource is employed to develop the empathetic dialogue dataset to improve affective understanding and empathetic expression.

\boldpara{Data Cleaning.}
To ensure the data quality, we implement several important data cleaning steps: 
(1) To process materials in non-textual formats such as images and PDFs, we employ \texttt{OLMOCR}\footnote{\url{https://olmocr.allenai.org/}} for accurate text recognition and conversion to text format; 
(2) We standardize the usage of Chinese and English punctuation and remove irrelevant content, including emojis, emoticons, and links, to filter out potential noise;
%
(3) To construct the empathetic dialogue data from resource \textbf{Type} \uppercase\expandafter{\romannumeral4}, we leverage LLMs to evaluate the reasonableness and relevance of counselor responses and filter out replies that lack substantive advice. 
For example, given the question ``I have recently experienced insomnia and feel anxious. What should I do?'', if the response is merely ``Everything will get better,'' it would be filtered due to the absence of practical advice.

\subsection{Psychological Reasoning Data Synthesis} 
\label{sec:psychological_data_synthesis}

\boldpara{Question Generation and Control.}
Following the data cleaning, we proceed to generate structured questions and corresponding answers from curated psychological textbooks and instructional materials (i.e., resource \textbf{Type} \uppercase\expandafter{\romannumeral1}).
Specifically, the source material is first segmented into multiple textual chunks, with each chunk designed to encapsulate the maximum amount of domain-specific content. 
Subsequently, we leverage LLMs to generate a set of different questions along with corresponding answers based on these text segments.
Meanwhile, for resource \textbf{Type} \uppercase\expandafter{\romannumeral3}, we use the similar way to generate QA pairs without segment-level contextual augmentation, aiming to supplement psychological subfields that are underrepresented or difficult to source from publicly accessible materials on the Internet.
Through these steps, we obtain approximately 200k generated QA pairs in total.

All generated QA pairs, together with data derived from resource \textbf{Type} \uppercase\expandafter{\romannumeral2}, are integrated into a unified QA pool containing approximately 210k entries.
%
This pool comprises resources \textbf{Type} \uppercase\expandafter{\romannumeral1}, \textbf{Type} \uppercase\expandafter{\romannumeral2} and \textbf{Type} \uppercase\expandafter{\romannumeral3}.
We implement a multi-stage quality control procedure to ensure the integrity and utility of the synthesized data.
Concretely, we use min-hash locality-sensitive hashing (LSH) to cluster similar questions and select the optimal one through LLM-based ranking.
Afterward, we prompt the LLM with few-shot examples to identify and filter out low-quality questions, specifically those exhibiting incomplete information, logical confusion, or unclear expression.
%
Finally, we invite 10 undergraduate and graduate students to conduct a semi-automatic validation procedure.
They identify recurring error patterns and apply batch-cleaning rules to eliminate redundant content and reduce potential noise in the dataset, ultimately retaining about 90k QA pairs.

\boldpara{Rationale Generation.} 
\label{rationale_generation}
We further generate detailed rationales for the aforementioned questions through CoT prompting, following a multi-step reasoning approach \cite{hsieh-etal-2023-distilling} to provide clear reasoning paths for model training. 
In detail, the CoT prompt guides the model to first comprehend the question, recognize relevant psychological concepts and knowledge, and decompose the problem into a sequence of analytical steps.
At each stage of this process, the model is required to articulate an intermediate rationale, ultimately generating a final answer derived from the accumulated reasoning.
Formally, given a CoT prompt $P$ and a QA pair $(q_{i}, {a}_{i})$, this procedure yields a rationale-augmented instance $(q_{i}, r_{i}, \hat{a}_{i})$, where $r_i$ denotes the reasoning path and $\hat{a}_i$ the model-predicted answer.
If the predicted answer aligns with the ground truth (i.e., $\hat{a}_i = a_i$), we regard the $r_{i}$ as a valid rationale.
In contrast, if the predicted answer is incorrect (i.e., $\hat{a}_i \neq a_i$), we guide the model to regenerate the rationale up to $T$ time.
Instances that fail to produce correct predictions after $T$ regeneration attempts are pruned from the final curated dataset.
After obtaining the initial rationale, we employ a self-supervised optimization strategy to iteratively refine both the prompt and the rationale with the goal of enhancing their clarity and reliability.
Specifically, for each instance $(P, q_i, r_i, \hat{a}_i)$, the prompt $P$ and rationale $r_i$ are jointly updated over multiple rounds, enabling the model to progressively improve its reasoning process.
Each round of optimization process consists of two sequential steps:
\begin{itemize}[leftmargin=*]
    \item \textbf{Prompt refinement}: We first guide the LLM to generate an improved candidate prompt $P^*_i$ from the current prompt, question, and rationale, represented as $P^*_i \leftarrow LLM(P, q_i, r_i)$, aiming to enhance the reasoning guidance.
    \item \textbf{Rationale revision}: Based on the candidate prompt $P^*_i$, the LLM subsequently generates a revised rationale along with its corresponding predicted answer, denoted as $(r^*_i, \hat{a}_{i}^{*}) \leftarrow LLM(P^*_i, q_{i})$.
\end{itemize}
If the $\hat{a}_{i}^{*}$ matches the ground truth $a_{i}$, we retain $P^*_{i}$ as an updated prompt and continue iteration based on the updated instance $(P^*_{i}, q_{i}, r^*_{i}, \hat{a}_{i}^{*})$.
Otherwise, the process reverts to the previous prompt-rationale pair to maintain alignment with correct reasoning paths.
We repeat this process for $R$ rounds ($R=3$ in this paper). 
After completing all iterations, we evaluate the rationales generated in each round for a given question and select the one that demonstrates the highest quality, denoted as $(P^*_{i}, q_{i}, r^*_{i}, \hat{a}_{i}^{*}=a_{i})$.
%
At this stage, we retain approximately 75k high-quality instances from the initial set of 90k pairs obtained in the previous step, by pruning instances whose generated rationales fail to support an exact-match final answer even after all regeneration rounds.

\boldpara{Question Selection.} 
\label{question_selection}
While the preceding steps yield high-quality data, not all instances exhibit sufficient complexity to facilitate effective reinforcement learning (RL).
To address this, in this stage, we implement a multi-LLM cross-selection strategy aimed at identifying and isolating the most challenging psychology-related samples from the constructed dataset for subsequent use in the reinforcement learning phase.
In detail, we employ three distinct LLMs (i.e., \texttt{Qwen}, \texttt{Llama}, and \texttt{Phi}) to independently generate responses for each question in the constructed psychological data.
Questions that receive incorrect responses from all three models are aggregated into a challenging subset with 19k instances that provide sophisticated scenarios in psychology.
This subset is intended to represent highly difficult instances with strong potential to enhance the model's reasoning capabilities through reinforcement learning.

\subsection{Empathetic Dialogue Synthesis}
\label{sec:empathetic_dialogue_synthesis}
In addition to psychological QA pairs, empathy is recognized as a core component of effective mental health support \cite{sorin2024large}. 
To this end, we incorporate empathetic expressions into the dialogue corpus derived from authentic resources to enhance its emotional richness and relevance to real-world psychological interactions.
Specifically, we refine these dialogue data through LLMs to achieve the following objectives.
We first enhance emotional resonance by incorporating empathetic expressions and supportive techniques (e.g., ``Hearing about your experience, I wish I could give you a warm hug.'').
Subsequently, we ensure that each dialogue provides evidence-based guidance that facilitates deeper understanding of users' issues, instead of limiting responses to surface-level empathy.
Finally, we deliver solution-oriented support by offering concrete coping strategies and practical steps that address the specific issues and challenges presented.
Through these steps, we ultimately obtain 73k high-quality dialogue data equipped with sufficient empathetic expressions.

\subsection{Data Split}
Leveraging the aforementioned pipelines, we curate a comprehensive dataset that comprises over 75k psychological questions with detailed rationales, among which 19k are identified as challenging samples through multi-LLM cross-selection.
The challenging subset is denoted as $\mathcal{D}_{pc}$ and the remaining data are denoted as $\mathcal{D}_{pr}$.
In parallel, the dataset contains over 73k empathetic dialogues engineered for contextually appropriate psychosocial interactions, denoted as $\mathcal{D}_{em}$.
To further enrich our training data, we additionally introduce multi-turn dialogues and knowledge-based QA from the PsycoLLM dataset \cite{hu2024psycollm}, denoted as $\mathcal{D}_{ps}$, and a refined set of 8k examination questions from the CPsyExam train set \cite{zhao-etal-2025-cpsyexam}, denoted as $\mathcal{D}_{cp}$.

Ultimately, the curated datasets are partitioned into two distinct subsets aligned with specialized training objectives.
One category, represented as $\mathcal{D}_{sft} = \mathcal{D}_{pr} \cup \mathcal{D}_{em} \cup \mathcal{D}_{ps}$, is designated for SFT.
The other category, denoted as $\mathcal{D}_{grpo} = \mathcal{D}_{pc} \cup \mathcal{D}_{cp}$, is reserved for GRPO.
Detailed prompts for the data synthesis pipeline are provided in the Appendix \ref{chap: appendix_prompts}.

\subsection{Model Training}
To enhance both reasoning capabilities and performance in empathy and expertise, we employ a hybrid training strategy.
\boldpara{Stage 1: Supervised Fine-Tuning.} 
We initialize our backbone model $\pi_{\theta}$ with \texttt{Qwen2.5-7B-Instruct} \cite{qwen2.5} and finetune it on $\mathcal{D}_{\text{sft}}$.
Formally, given a query $x$, the model is trained to generate a coherent rationale $r$ followed by a corresponding answer $a$, where the complete output is denoted as $y = [r; a]$.
We optimize model parameters $\theta$ by minimizing the standard negative log-likelihood loss:
\begin{equation}
\resizebox{0.85\linewidth}{!}{$ 
    \displaystyle 
    \mathcal{L}(\theta) = - \mathbb{E}_{(x, y) \sim \mathcal{D}_{\text{sft}}} \left[ \sum_{t=1}^{T} \log P(y_t \mid x, y_{<t}; \theta) \right]
$}
\end{equation}

\boldpara{Stage 2: Group Relative Policy Optimization.} 
To further refine psychological reasoning proficiency, we employ GRPO \cite{shao2024deepseekmath} on $\mathcal{D}_{\text{grpo}}$.
We design a composite reward function $R(y, y^*) = R_{\text{fmt}} + R_{\text{acc}}$ to guide policy learning, where $y^*$ denotes the ground truth.
\begin{itemize}[leftmargin=*]
    \item \textbf{Format reward ($R_{\text{fmt}}$)}: 
    We enforce strict formatting constraints.
    The reasoning process must be encapsulated within \texttt{<think>} and \texttt{</think>} tags, followed by the final answer.
    We assign a scalar reward $R_{\text{fmt}} = +1.25$ for structurally parsable outputs and $-1$ otherwise.
    \item \textbf{Accuracy reward ($R_{\text{acc}}$)}: 
    We formulate the answer matching as a set comparison task.
    Specifically, we parse the predicted answer $\hat{a}$ and the ground truth $a$ into sets of discrete options.
    To encourage precise reasoning alignment while penalizing hallucinations or omissions, we employ a partial-credit mechanism based on the overlap between the prediction and the ground truth:
\begin{equation}
    R_{\text{acc}} = 
    \begin{cases}
    +1, & \text{if } \hat{a} = a \\
    \frac{|\hat{a} \cap a|}{|a|}, & \text{if } \hat{a} \subset a \land a \neq \emptyset \\
    -1, & \text{otherwise}
    \end{cases}
    \end{equation}
\end{itemize}
By integrating these logical and structural signals, the model learns to generate well-organized reasoning processes while rewarding partial credit for incomplete but valid answers, leading to better performance in psychological tasks.

\begin{table*}[ht]
  \centering
    \small
     \setlength{\tabcolsep}{1mm} 
\resizebox{0.99\textwidth}{!}{
  \begin{tabular}{l|c|ccccccccc|cc|ccc}
    \toprule

    \multirow{2}{*}{\textbf{Model}} & \multirow{2}{*}{\textbf{Param.}}
      & \multicolumn{3}{c|}{\textbf{Case}}
      & \multicolumn{3}{c|}{\textbf{Moral}}
      & \multicolumn{3}{c|}{\textbf{Theory}}
      & \multicolumn{2}{c|}{\multirow[c]{2}{*}{\textbf{Avg.}}}
      & \multicolumn{3}{c}{\textbf{Case (QA)}} \\
      
      & & \textbf{SMCQ}
      & \multicolumn{2}{c|}{\textbf{MMCQ}}
      & \textbf{SMCQ}
      & \multicolumn{2}{c|}{\textbf{MMCQ}}
      & \textbf{SMCQ}
      & \multicolumn{2}{c|}{\textbf{MMCQ}}
      & \multicolumn{2}{c|}{\textbf{}}
      & \textbf{R-1} & \textbf{R-L} & \textbf{B-4} \\

    \midrule

\rowcolor{gray!60}
\multicolumn{16}{c}{\textit{\textbf{General LLMs}}}   \\

MiniCPM4-8B      & 8B & 50.00 & 28.59 &  \underline{43.64} & 81.58 &  50.63 & \underline{58.23} &  65.62 & 34.06 & \underline{43.00} & 51.75 & (\underline{57.01}) & 23.05 & 12.90 & 1.35 
\\
Qwen2.5-7B      & 7B & 47.57 & 31.64 &  \underline{47.49} & 87.83 &  59.50 & \underline{71.02} &  78.46 & 42.45 & \underline{55.17} & 57.91 & (\underline{64.59}) & 20.94 & 11.28 & 1.28 
\\
Qwen2.5-14B      & 14B & 47.13 & 41.10 &  \underline{55.93} & 89.81 &  63.93 & \underline{73.60} &  80.32 & 50.16 & \underline{61.26} & 62.08 & (\underline{68.01}) & 22.69 & \cellcolor[HTML]{FFFFD4}13.93 & \cellcolor[HTML]{FFFFD4}1.53
\\
Qwen2.5-72B      & 72B & 46.91 & 40.34 &  \underline{53.11} & 90.79 &  \cellcolor[HTML]{FFE4CF}70.25 & \underline{78.48} &  82.63 & 47.63 & \underline{59.74} & 63.09 & (\underline{68.61}) & 21.43 & 12.02 & 1.16 
\\

\rowcolor{gray!60}
\multicolumn{16}{c}{\textit{\textbf{Reasoning-Augmented LLMs}}}   \\

DeepSeek-R1        & 671B & \cellcolor[HTML]{FFCCC9}79.25 & \cellcolor[HTML]{FFE4CF}44.25 & \cellcolor[HTML]{FFE4CF}\underline{60.86} & \cellcolor[HTML]{FFCCC9}95.39 &  68.99 & \underline{77.95} &  \cellcolor[HTML]{FFCCC9}92.19 & \cellcolor[HTML]{FFE4CF}57.60 & \cellcolor[HTML]{FFE4CF}\underline{69.41} & \cellcolor[HTML]{FFE4CF}72.95 & \cellcolor[HTML]{FFCCC9}(\underline{79.18}) & 17.65 & 9.19 & 0.94 
\\ 
DeepSeek-R1-70B  & 70B & 56.30 & 30.72 &  \underline{46.95} & 88.16 &  52.53 & \underline{65.66} &  68.01 & 25.64 & \underline{45.63} & 53.56 & (\underline{61.79}) & 22.77 & 13.23 & 1.16 
\\ 
QwQ-32B                  & 32B & 56.51 & 23.35 &  \underline{41.27} & 88.82 &  41.14 & \underline{53.06} &  82.12 & 32.69 & \underline{49.90}  & 54.11 & (\underline{61.95}) & 18.39 & 7.48 & 0.84 
\\ 
Qwen3-30B-A3B          & 30B & 59.65 & 31.51 &  \underline{47.28} & 91.45 &  55.06 & \underline{65.66} &  80.75 & 47.45 & \underline{59.25} & 60.98 & (\underline{67.34}) & 20.53  & 12.06 & 1.18 
\\
Qwen3-235B-A22B          & 235B & \cellcolor[HTML]{FFE4CF}68.58 & \cellcolor[HTML]{FFFFD4}41.91 &  \cellcolor[HTML]{FFFFD4}\underline{57.24} & \cellcolor[HTML]{FFE4CF}93.42 &  69.62 & \cellcolor[HTML]{FFE4CF}\underline{78.90} &  \cellcolor[HTML]{FFE4CF}88.36 & \cellcolor[HTML]{FFFFD4}56.70 & \cellcolor[HTML]{FFFFD4}\underline{68.64} & \cellcolor[HTML]{FFFFD4}69.77 & \cellcolor[HTML]{FFFFD4}(\underline{75.86}) & 18.96 & 11.14 & 1.11 
\\
Magistral-Small  & 24B & 56.58 & 33.26 &  \underline{49.11} & 82.89 & 53.80 & \underline{67.99} &  70.10 & 37.76 & \underline{52.35} & 55.73 & (\underline{63.17}) & 22.90 & 11.97 & 1.21  
\\ 

\rowcolor{gray!60}
\multicolumn{16}{c}{\textit{\textbf{Closed-Source LLMs}}}   \\

Claude3.7-Sonnet   & UNK & 63.39 & 19.40 & \underline{34.23} & 90.13 &  60.13 & \underline{70.04} & 76.73 & 37.37 & \underline{48.99} & 57.86 & (\underline{63.92}) & 21.59 & 11.11 & 1.23 
\\ 

Gemini1.5-Pro   & UNK & 61.04 & 35.57 &  \underline{49.87} & 84.87 &  62.03 & \underline{70.62} & 80.84 & 43.22 & \underline{53.44} & 61.26 & (\underline{66.78}) & 21.63 & 10.93 & 1.06 
\\ 

GPT-4o    & UNK & \cellcolor[HTML]{FFFFD4}65.63 & 13.67 &  \underline{34.53} & 88.15 & 33.54 & \underline{54.79} &  74.65 & 24.10 & \underline{45.07} & 49.96 & (\underline{60.47}) & \cellcolor[HTML]{FFFFD4}23.45 & 12.75 & 1.18  
\\

\rowcolor{gray!60}
\multicolumn{16}{c}{\textit{\textbf{Psychological LLMs}}}   \\

CPsyCounX                & 7B & 40.87 & 16.91 &  \underline{32.90} & 75.17 &  36.08 & \underline{54.85} &  54.78 & 19.03 & \underline{38.90} & 40.47 & (\underline{49.58}) & 22.83 & 11.94 & 1.48
\\ 

EmoLLM          & 7B & 46.93 & 21.87 &  \underline{40.02} & 84.21 &  34.17 & \underline{51.05} &  71.72 & 26.18 & \underline{44.49} & 47.51 & (\underline{56.40}) & 22.15 & 11.69 & 1.20 
\\

PsycoLLM                 & 14B & 55.58 & 35.07 &  \underline{42.89} & 88.81 &  69.62 & \underline{74.20} &  72.63 & 48.59 & \underline{54.12} & 61.72 & (\underline{64.71}) & \cellcolor[HTML]{FFE4CF}24.45 & \cellcolor[HTML]{FFCCC9}17.45 & \cellcolor[HTML]{FFE4CF}2.04
\\ 

PsyDT                 & 7B & 35.56 & 35.20 &  \underline{50.14} & 86.33 &  \cellcolor[HTML]{FFFFD4}69.70 & \cellcolor[HTML]{FFFFD4}\underline{78.66} &  80.70 & 52.72 & \underline{62.26} & 60.04 & (\underline{65.61}) & 20.65 & 13.41 & 1.16
\\ 

Psyche-R1           & 7B & 63.31 & \cellcolor[HTML]{FFCCC9}56.26 & \cellcolor[HTML]{FFCCC9}\underline{66.21} & \cellcolor[HTML]{FFFFD4}92.76 &  \cellcolor[HTML]{FFCCC9}79.62 & \cellcolor[HTML]{FFCCC9}\underline{82.54} &  \cellcolor[HTML]{FFFFD4}87.70 & \cellcolor[HTML]{FFCCC9}66.54 & \cellcolor[HTML]{FFCCC9}\underline{73.34} & \cellcolor[HTML]{FFCCC9}74.37 & \cellcolor[HTML]{FFE4CF}(\underline{77.64}) & \cellcolor[HTML]{FFCCC9}27.31 & \cellcolor[HTML]{FFE4CF}15.33 & \cellcolor[HTML]{FFCCC9}2.40
\\
    \bottomrule
  \end{tabular}
  }
  \caption{
Comparison of different models on the PCEB, where Case, Moral, Theory, and Case (QA) are case analysis, theoretical proficiency, professional ethics, and case-based QA.
The average value represents the average of the standard accuracy rates, and values in parentheses denotes the mean of the standard accuracy for SMCQ and the elastic accuracy for MMCQ.
Results colored in red, orange, and yellow demote the best, second-best and third-best.
  }
\label{tab:model_comparison_sorted}
\vspace{-1em}
\end{table*}

\section{Experiments}
\subsection{Experimental Setting}
\boldpara{Baselines.} 
To ensure a comprehensive analysis, we selected 17 representative LLMs, categorized as follows: (1) \textbf{General LLMs,} which exhibit excellent performance across general tasks, but lack explicit reasoning capabilities.
(2) \textbf{Reasoning augmented LLMs,} which possess explicit reasoning capabilities.
(3) \textbf{Closed-source LLMs,} which typically represent the state-of-the-art performance.
(4) \textbf{Psychological LLMs,} which have been fine-tuned on psychological datasets.
A detailed summary of all models is presented in Appendix \ref{chap: appendix_baseline}.

\boldpara{Benchmarks and Evaluation Metrics.}
We conduct comprehensive evaluations on two professional psychological benchmarks:
\begin{itemize}[leftmargin=*]
    \item \textbf{Psychological counselor examination benchmark (PCEB)} \cite{hu2024psycollm}: this consists of 3,863 multiple-choice questions (MCQ) and 100 open-ended case analysis items, curated from the official National Psychological Counselor Examination in China.
    \item \textbf{CPsyExam test set} \cite{zhao-etal-2025-cpsyexam}: this includes 4,102 questions spanning 39 distinct psychological subfields. We evaluate under both zero-shot and five-shot settings, ensuring consistency by using identical exemplars across all evaluated models in the latter setting.
\end{itemize}

Note that MCQ comprises two types of questions: MCQ with only a single correct option (SMCQ), and MCQ with multiple correct options (MMCQ).
For MCQ, we adopt metrics introduced in \citet{hu2024psycollm}, including \textbf{standard accuracy}, which requires predictions to exactly match the ground truth, and \textbf{elastic accuracy}, which gives partial credit when predictions are a subset of the correct answers.
For open-ended questions, we utilize the existing text generation metrics, including \textbf{ROUGE-1 (R-1)}, \textbf{ROUGE-L (R-L)} \cite{lin-2004-rouge}, and \textbf{BLEU-4 (B-4)} \cite{papineni-etal-2002-bleu}. 

\begin{table*}[ht]

\small
  \centering
\resizebox{0.98\textwidth}{!}{

\begin{tabular}{l|c|cccc|cccc|c}
    \toprule

    \multirow{3.4}{*}{\textbf{Model}} & \multirow{3.4}{*}{\textbf{Param.}}
      & \multicolumn{4}{c|}{\textbf{Zero-Shot}}
      & \multicolumn{4}{c|}{\textbf{Five-Shot}}
      & {\multirow[c]{3.4}{*}{\textbf{Avg.}}}
      \\

\cline{3-6} \cline{7-10} 
      
  & & \multicolumn{2}{c}{\textbf{Knowledge}}
    & \multicolumn{2}{c|}{\textbf{Case}}
    & \multicolumn{2}{c}{\textbf{Knowledge}}
    & \multicolumn{2}{c|}{\textbf{Case}}
    & \\ 
  & & \textbf{SMCQ} & \textbf{MMCQ}
    & \textbf{SMCQ} & \textbf{MMCQ}
    & \textbf{SMCQ} & \textbf{MMCQ}
    & \textbf{SMCQ} & \textbf{MMCQ}
    & \\
      
  \midrule


\rowcolor{gray!60}
\multicolumn{11}{c}{\textit{\textbf{General LLMs}}}   \\

MiniCPM4-8B & 8B & 69.58 & 41.74 & 57.33 & 37.00 & 68.50 & 42.77 & 54.67 & 38.00 & 60.46 \\ 
Qwen2.5-7B & 7B & 76.99 & 43.66 & 68.67 & 44.50 & 78.63 & 42.00 & 68.67 & 40.50 & 67.37 \\ 
Qwen2.5-14B & 14B & 81.39 & 49.30 & 72.00 & 48.50 & 82.42 & 54.29 & 71.00 & 48.00 & 71.84 \\
Qwen2.5-72B & 72B & \cellcolor[HTML]{FFE4CF}84.61 & \cellcolor[HTML]{FFFFD4}52.75 & \cellcolor[HTML]{FFE4CF}73.50 & \cellcolor[HTML]{FFE4CF}54.50 & \cellcolor[HTML]{FFE4CF}86.64 & \cellcolor[HTML]{FFE4CF}63.77 & \cellcolor[HTML]{FFFFD4}75.33 & \cellcolor[HTML]{FFE4CF}55.00 & \cellcolor[HTML]{FFE4CF}74.98 \\

\rowcolor{gray!60}
\multicolumn{11}{c}{\textit{\textbf{Reasoning-Augmented LLMs}}}   \\

DeepSeek-R1 & 671B & \cellcolor[HTML]{FFCCC9}87.49 & \cellcolor[HTML]{FFE4CF}56.98 & \cellcolor[HTML]{FFCCC9}76.83 & \cellcolor[HTML]{FFCCC9}59.00 & \cellcolor[HTML]{FFCCC9}88.78 &  \cellcolor[HTML]{FFCCC9}66.58 & \cellcolor[HTML]{FFCCC9}77.30  & \cellcolor[HTML]{FFCCC9}61.50 & \cellcolor[HTML]{FFCCC9}78.28 \\ 
DeepSeek-R1-70B & 70B & 76.48 & 22.80 & 61.81 & 19.17 & 76.89 & 40.99 & 62.70 & 37.95 & 60.57 \\ 

\rowcolor{gray!60}
\multicolumn{11}{c}{\textit{\textbf{Closed-Source LLMs}}}   \\

Gemini1.5-Pro & UNK & 82.08 & 40.59 & 68.33 & 43.00 & \cellcolor[HTML]{FFFFD4}83.93 & 53.65 & 71.00 & 45.00 & 69.66 \\ 
GPT-4o & UNK & 80.70 & 30.73 & 66.33 & 28.00 & 81.82 & 54.80 & 68.67 & 51.50 & 65.79 \\

\rowcolor{gray!60}
\multicolumn{11}{c}{\textit{\textbf{Psychological LLMs}}}   \\

CPsyCounX & 7B & 57.56 & 22.41 & 46.33 & 31.00 & 63.46 & 21.77 & 50.67 & 23.50 & 47.44 \\ 
EmoLLM & 7B & 78.41 & 45.33 & \cellcolor[HTML]{FFFFD4}72.50 & 48.00 & 79.92 & 36.88 & 74.17 & 39.50 & 69.32 \\
PsycoLLM & 14B & 78.33 & 51.98 &  65.33 & 42.00 & 78.63 & 50.45 & 65.57 & 36.00 & 69.20 \\ 
PsyDT & 7B & 80.83 & 48.91 &  69.67 & 41.50 & 81.13 & 40.97 & 68.33 & 40.00 & 70.71 \\ 

Psyche-R1 & 7B & \cellcolor[HTML]{FFFFD4}82.72 & \cellcolor[HTML]{FFCCC9}61.59 & 70.50 & \cellcolor[HTML]{FFFFD4}49.50  & 83.45 & \cellcolor[HTML]{FFFFD4}61.46 & \cellcolor[HTML]{FFE4CF}76.17 & \cellcolor[HTML]{FFFFD4}52.00 & \cellcolor[HTML]{FFFFD4}74.90 \\ 
    \bottomrule
  \end{tabular}
  }
  \caption{  
Comparisons of different models on the CPsyExam test set. The average represents the overall zero-shot accuracy. The first, second, and third-best results are marked in red, orange, and yellow, respectively.
  }
  \label{tab:performance1}
  \vspace{-1em}
\end{table*}

\begin{table}[ht]
\centering
\renewcommand{\arraystretch}{1.1}
\setlength{\tabcolsep}{1.2mm}
\resizebox{0.48\textwidth}{!}{
\begin{tabular}{l|cc|cc|cc}

\toprule

\textbf{Model} 
& \textbf{Case} 
& $\Delta$ 
& \textbf{Moral} 
& $\Delta$ 
& \textbf{Theory} 
& $\Delta$  
\\

\midrule
Base        & 38.97 & \cellcolor{gray!10}- & 73.39 & \cellcolor{gray!10}- & 63.83 & \cellcolor{gray!10}- \\
\midrule

\rowcolor{gray!60} 
\multicolumn{7}{l}{\textit{\textbf{Ablation study on training stage}}}   \\

+GRPO         & 36.69 &  \cellcolor{red!10}-5.85\% & 77.74 &  \cellcolor{PineGreen!10}5.93\% & 73.34 &  \cellcolor{PineGreen!15}14.90\% \\

+SFT        & 48.51 &  \cellcolor{PineGreen!25}24.48\% & 83.82 &  \cellcolor{PineGreen!15}14.21\% & 73.44 &  \cellcolor{PineGreen!15}15.06\% \\

+SFT+GRPO     & 67.07 &  \cellcolor{PineGreen!70}72.11\% & 86.06 &  \cellcolor{PineGreen!18}17.26\% & 79.10 &  \cellcolor{PineGreen!25}23.92\% \\

\midrule

\rowcolor{gray!60}
\multicolumn{7}{l}{\textit{\textbf{Ablation study on rationale optimization}}}   \\

+QA         & 48.27 &  \cellcolor{PineGreen!25}23.86\% & 81.89 &  \cellcolor{PineGreen!12}11.58\% & 71.22 &  \cellcolor{PineGreen!12}11.58\% \\

+Rat.       & 55.25 &  \cellcolor{PineGreen!45}41.78\% & 85.14 &  \cellcolor{PineGreen!16}16.01\% & 75.55 &  \cellcolor{PineGreen!20}18.36\% \\

+Rat.+Iter. & 67.07 &  \cellcolor{PineGreen!70}72.11\% & 86.06 &  \cellcolor{PineGreen!18}17.26\% & 79.10 &  \cellcolor{PineGreen!25}23.92\% \\

\bottomrule
\end{tabular}
}
\caption{
Ablation study evaluating the effects of training stages (SFT and GRPO) and the contributions of rationale component (Rat.) and iterative prompt–rationale optimization (Iter.).
}
\label{tab:ablation_training_and_rationale}
\vspace{-1em}
\end{table}

\subsection{Overall Results} 
\boldpara{Results on the PCEB.}
To evaluate the performance of different models, we present the results on the PCEB in Table \ref{tab:model_comparison_sorted}. These results reveal several key observations.
First, \textit{Psyche-R1} exhibits strong performance across evaluation tasks in both MCQ and subjective questions.
This demonstrates the effectiveness of our proposed dataset and training strategy in simultaneously enhancing psychological reasoning and text generation capabilities for psychological tasks.
Second, while \texttt{DeepSeek-R1} excels in MCQ, its performance in subjective questions is notably limited.
This performance disparity can be attributed to its training methodology, which employs RL on datasets primarily consisting of mathematical and coding tasks with deterministic solutions.
Although this approach strengthens logical reasoning, it appears to bias the model towards single-answer patterns, thereby limiting its capability to generate diverse and nuanced responses in open-ended psychological assessments.
Third, existing psychological LLMs (e.g., \texttt{CPsyCounX} and \texttt{EmoLLM}) achieve strong performance in subjective questions while demonstrating limited abilities in MCQ.
This imbalanced performance stems from their reliance on training exclusively on counseling dialogues or empathetic conversations, which constrains their capabilities to develop comprehensive competencies.
Fourth, closed-source models such as \texttt{GPT-4o} and \texttt{Claude3.7-Sonnet} demonstrate relatively weaker performance, which may be attributed to limited Chinese representation in their training corpora.

\boldpara{Results on the CPsyExam Test Set.}
To further explore model performance, we present the results on the CPsyExam test set in Table \ref{tab:performance1}. 
Similar to the trends observed in previous experiments, both \textit{Psyche-R1} and \texttt{DeepSeek-R1} demonstrate superior performance. 
Across these models, psychological LLMs consistently achieve higher accuracy in SMCQ than in MMCQ, as the latter requires exhaustive evaluation of all options, demanding more comprehensive domain-specific knowledge and reasoning capabilities.
Under the five-shot setting, most models exhibit substantial improvements in MMCQ (e.g., \texttt{PsyDT} achieves a 47.64\% improvement in knowledge-type MMCQ). 
This observation aligns with existing studies, which demonstrate that well-designed few-shot examples can effectively enhance model performance in certain tasks.
In contrast, \texttt{DeepSeek-R1} exhibits a performance decline under the five-shot compared to its zero-shot setting, suggesting that few-shot prompting may interfere with its inherent reasoning capability aligning with existing findings \cite{guo2025deepseek}.

\subsection{Ablation Study}

In this section, we conduct a comprehensive ablation study, evaluating model performance by standard accuracy on the PCEB.

\boldpara{Effect of SFT and GRPO.} 
We investigate the effect of SFT and GRPO, with results shown in Table \ref{tab:ablation_training_and_rationale}.
We can observe that SFT substantially improves performance across the three task categories by leveraging our dataset of empathetic dialogues and rationale-augmented psychological questions.
However, applying GRPO without prior SFT results in performance degradation on SMCQ case tasks, as the base model lacks sufficient domain knowledge and empathy, which are critical for reasoning in case-based questions, leading to unstable training dynamics.
When combining SFT with GRPO training yields further gains, particularly on case-based tasks, as challenging samples identified via multi-LLM cross-selection promote deeper reasoning and contextual understanding.

\boldpara{Effect of the Rationale and Iterative Optimization.}
We explore the contributions of rationales and iterative prompt–rationale optimization, with results presented in Table \ref{tab:ablation_training_and_rationale}.
Note that \textbf{+QA} is the model trained solely on question–answer pairs without incorporating detailed rationales.
Compared with the base model, training with our proposed dataset (i.e., \textbf{+QA}, \textbf{+Rat.} and \textbf{+Rat.+Iter.}) leads to consistent performance improvements, demonstrating the effectiveness of the curated data.
Integrating rationale-augmented data substantially enhances performance over training with option-only answers, indicating that rationales provide valuable intermediate reasoning signals that facilitate learning.
Furthermore, applying iterative prompt–rationale optimization (i.e., \textbf{+Rat.+Iter.}) yields further gains, confirming that progressively refined rationales contribute to better supervision and more robust model reasoning.

\begin{figure}[t]
\centering
\includegraphics[width=0.97\columnwidth]{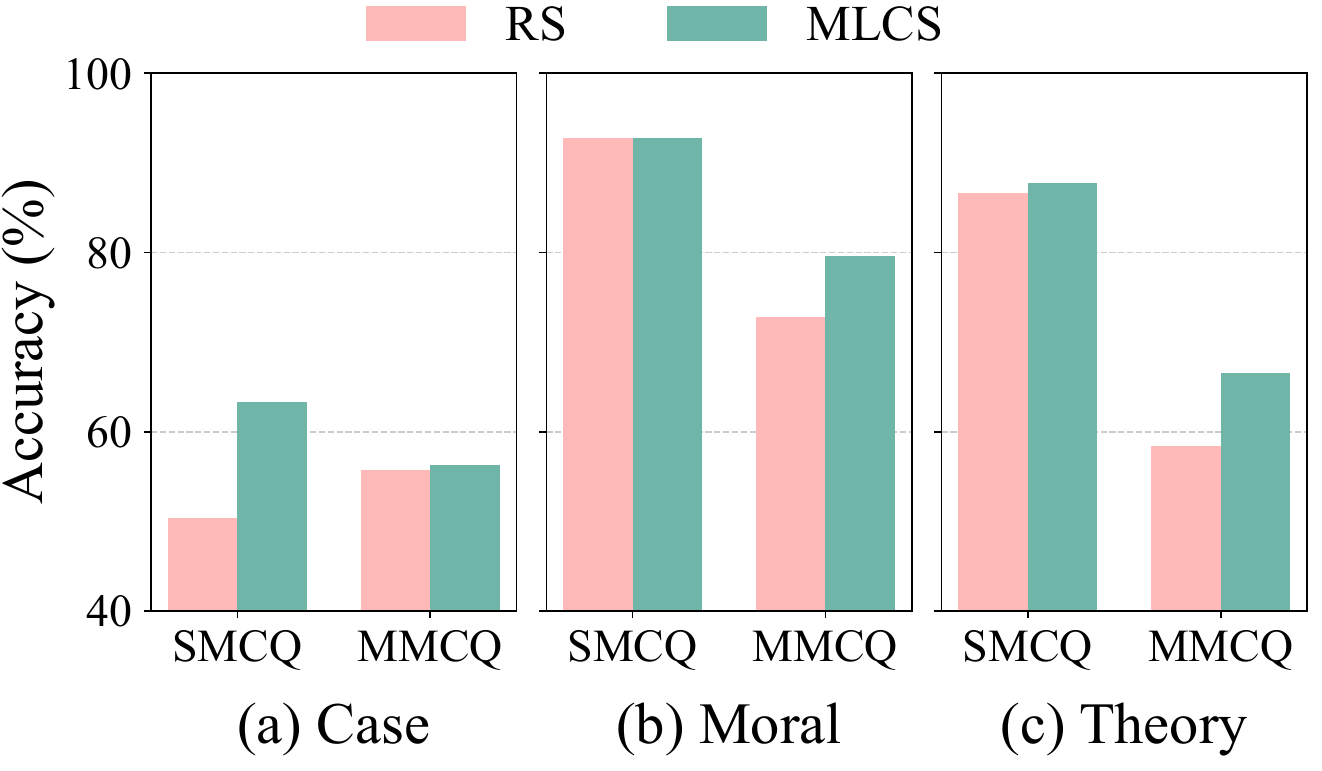} 
\vspace{-1em}
\caption{
Comparison of performance using challenging question selection methods, including multi-LLM cross-selection (MLCS) and random selection (RS).
}
\label{methods_comparison}
\vspace{-1em}
\end{figure}

\boldpara{Effect of the Question Selection.}
We further examine the effect of question selection by comparing multi-LLM cross-selection (MLCS) with random selection (RS), as illustrated in Figure \ref{methods_comparison}.
The comparison between MLCS and RS demonstrates that leveraging multiple LLMs for selecting challenging instances yields markedly superior outcomes across all these tasks.
This finding confirms that our selection mechanism effectively selects high-quality training data for GRPO, which is instrumental in enhancing reasoning and generalization within the psychological domain.

\subsection{Discussion}
\boldpara{Effect of Datasets.}
We evaluate the model performance across different combinations of subsets, with results presented in Table \ref{tab:ablation_dataset}.
It is observed that fine-tuning with either psychological reasoning data (PRD) or empathetic dialogues (ED) in isolation delivered marginal improvements in task performance, and in some cases, led to a slight decline in overall accuracy. 
In contrast, the combination of PRD and ED achieves substantial improvements across these tasks, highlighting the quality and comprehensiveness of our proposed data synthesis pipeline. 
This result demonstrates that integrating domain-specific knowledge with emotional understanding enhances psychological reasoning. 
Moreover, the incorporation of additional public datasets (APD) leads to further performance improvements.

\begin{table}[t]
 
 \centering
 \small
\setlength{\tabcolsep}{1mm}
  \resizebox{0.48\textwidth}{!}{
\begin{tabular}{l|cccccc|c}
  
  \toprule
  \multirow{2}{*}{\textbf{Settings}} 
  & \multicolumn{2}{c|}{\textbf{Case}} 
  & \multicolumn{2}{c|}{\textbf{Moral}} 
  & \multicolumn{2}{c|}{\textbf{Theory}} 
  & \multicolumn{1}{c}{\multirow[c]{2}{*}{\textbf{Avg.}}}
  \\

  & \textbf{SMCQ} 
  & \multicolumn{1}{c|}{\textbf{MMCQ}}
  & \textbf{SMCQ} 
  & \multicolumn{1}{c|}{\textbf{MMCQ}}
  & \textbf{SMCQ} 
  & \textbf{MMCQ} & \\

  \midrule

Base  & 47.57 & 31.64 & 87.83 & 59.50 & 78.46 & 42.45 & 57.91 \\
\midrule
+ED    & 35.77 & 29.74 & 70.00 & 60.90 & 65.84 & 44.40 & 51.11 \\ 
+PRD    & 37.94 & 35.45 & 91.45 & 51.27 & 80.47 & 37.53 & 55.69 \\
+ED+PRD    & 61.71 & 53.56 & 92.72 & 76.58 & 86.13 & 68.16 & 73.14 \\
  \midrule
+ED+PRD+APD  & 63.31 & 56.26 & 92.76 & 79.62 & 87.70 & 66.54 & 74.37 \\
  \bottomrule
 \end{tabular}
 }

 \caption{
 Effect of different dataset compositions, including empathetic dialogues (ED, i.e., $\mathcal{D}_{em}$), psychological reasoning data (PRD, i.e., $\mathcal{D}_{pc} \cup \mathcal{D}_{pr}$), and additional public datasets (APD, i.e., $\mathcal{D}_{ps} \cup \mathcal{D}_{cp}$).
 }
 \label{tab:ablation_dataset}
 \vspace{-1em}
\end{table}

\begin{table}[t]

 \centering
 \small
 \setlength{\tabcolsep}{1mm}
 \resizebox{0.42\textwidth}{!}{

\begin{tabular}{l|cccc}

  \toprule
  \textbf{Model} & \textbf{EmoE.} & \textbf{CogE.} & \textbf{Con.} &\textbf{Sta.}\\
  \midrule
  Qwen2.5-7B-Instruct & 1.52  & 2.00  & 2.36 & 1.72 \\

  CPsyCounX & 1.73  & 2.05  & 2.15  & 1.96 \\
  EmoLLM & 1.86  & 2.44  & \textbf{2.84} & \textbf{2.34} \\
  PsycoLLM & 1.97  & 2.27  & 2.41  & 2.10 \\
  PsyDT & 2.21  & 2.46  & 2.36 & \textbf{2.34} \\
\midrule
  Psyche-R1 & \textbf{2.33}  & \textbf{2.69}  & 2.78  & 2.11 \\
  \bottomrule
 \end{tabular}
 }

 \caption{Comparisons of psychological LLMs on PsyDT test set. The evaluation metrics comprise: emotional empathy (EmoE.), cognitive empathy (CogE.), conversation strategy (Con.), and state and attitude (Sta.).}
 \label{tab:qualitative_comparison}
\vspace{-1.3em}
\end{table}

\begin{figure}[t]
\centering
\includegraphics[width=\columnwidth]{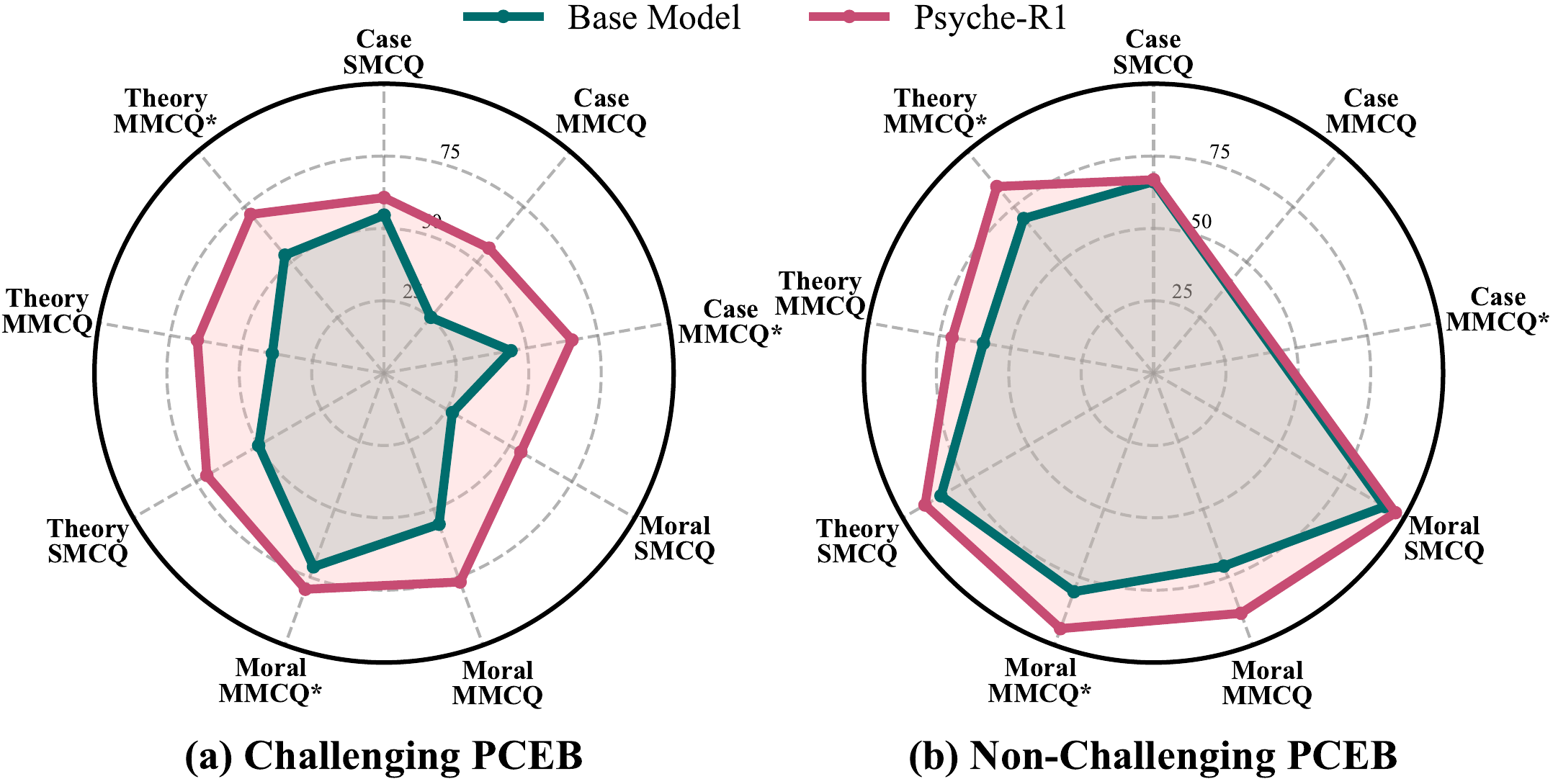}
\vspace{-1em}
\caption{
Comparison of model performance on (a) Challenging and (b) Non-Challenging subsets of the PCEB. An asterisk(*) indicates elastic accuracy, while the remaining metrics represent standard accuracy.
}
\label{fig4}
\vspace{-1em}
\end{figure}

\boldpara{Performance on Counseling Tasks.}
Beyond examination tasks, we evaluate the performance of \textit{Psyche-R1} on counseling tasks and compare it with its base model and several outstanding psychological LLMs.
Following the method of \texttt{PsyDT} \cite{xie-etal-2025-psydt} but constrained by limited resources, we randomly sample 200 instances from its test set and employ \texttt{GPT-4o} as the evaluator.
As shown in Table \ref{tab:qualitative_comparison}, \textit{Psyche-R1} achieves notable improvements compared to its base model, demonstrating its capability in counseling tasks that demand emotional empathy, cognitive empathy and so on.
This excellent performance stems from the synergistic interplay between two crucial elements: the empathetic dialogues, which directly improve counseling effectiveness, and advanced reasoning mechanisms, which enable a deeper understanding of questions, thereby yielding more accurate and emotionally informed responses within relevant contexts.

\boldpara{Error Analysis.}
To assess the impact of our approach across varying difficulty levels, we divide the PCEB into (a) challenging and (b) non-challenging subsets following the question selection method described in \S\ref{question_selection}.
The results are presented in Figure \ref{fig4}.
It is observed that \textit{Psyche-R1} demonstrates consistent improvements across these tasks in both subsets.
For the non-challenging one, performance gains are primarily concentrated in the theory and moral dimensions, reflecting the model's proficiency in handling foundational psychological concepts.
Notably, our model exhibits more pronounced improvements on the challenging subset.
This observation can be attributed to the synergistic effect of our data generation pipeline and hybrid training strategy.

\section{Related Work}
\subsection{LLMs for Psychology} 
The success of LLMs has spurred interest in developing LLM-driven mental health applications \cite{demszky2023using, xiao2025mentrasuiteposttraininglargelanguage, shen2026psychethicsbenchevaluatinglargelanguage}. 
Early research focused primarily on improving the accessibility of mental health services. 
Research in this phase primarily concentrated on two directions:
One direction involves leveraging NLP techniques for emotion recognition to enable automated detection of depression \cite{8882493} and suicidal ideation \cite{lee-etal-2020-cross}. 
The other focuses on constructing empathetic dialogue systems by fine-tuning LLMs on single-turn \cite{lai2023supporting} or multi-turn \cite{qiu-etal-2024-smile} dialogue data to enhance their abilities in affective understanding and emotional support \cite{2024EmoLLM, xie-etal-2025-psydt}. 
As research progressed, researchers began to explore more diverse mental health applications.
Some studies have transformed traditional psychometric tools (e.g., psychological scales) into interactive systems to improve user engagement \cite{kuribayashi-etal-2024-psychometric, yang-etal-2024-psychogat, xu-etal-2025-multiagentesc, Hu_Wang_Xie_Li_Ma_Guo_2026}. 
Another line of research has focused on the specialized demands of the psychological domain, developing professional-grounded mental health applications based on established psychological therapies \cite{lee-etal-2024-cactus, 10821773} or concepts \cite{zhang2025conceptpsy}.

\subsection{LLM Reasoning} 
In recent years, techniques such as CoT prompting \cite{NEURIPS2022_9d560961, hsieh-etal-2023-distilling} and STaR \cite{NEURIPS2022_639a9a17} have significantly advanced the development of LLM reasoning.
Building upon this foundation, researchers have explored more sophisticated reasoning architectures \cite{zhang2026semanticawarelogicalreasoningsemiotic, zhang2026logicalphasetransitionsunderstanding, shi2026reasoningtreesimprovingretrievalaugmented}.
For instance, Tree of Thoughts \cite{NEURIPS2023_271db992} enables systematic exploration of multiple reasoning paths with self-evaluation, while PAL \cite{pmlr-v202-gao23f} integrates reasoning with external tools through program generation.
These approaches further enhance model performance in handling complex tasks.
A new breakthrough was marked by the release of reasoning LLMs such as \texttt{OpenAI o1} \cite{openai2024openaio1card} and \texttt{DeepSeek-R1} \cite{guo2025deepseek}. 
These models, which are trained through reinforcement learning with reasoning techniques to enhance reasoning capabilities, demonstrate exceptional performance in mathematical and coding tasks \cite{comanici2025gemini, yang2025qwen3technicalreport}.
Motivated by these advances, researchers have employed advanced RL algorithms \cite{shao2024deepseekmath, yu2025dapo, zhou2026lookinwardexploreoutward} to further extend reasoning capabilities to domain-specific applications, including medicine \cite{liu2025beyond} and finance \cite{zhu2025dianjin}.
However, within the field of psychology, limited research has investigated the utility of reasoning.
To our knowledge, \textit{Psyche-R1} is the first psychological LLM that unifies empathy, domain-specific expertise, and reasoning capabilities.

\section{Conclusion}
In this paper, we propose \logopsyche\textit{Psyche-R1}, the first Chinese psychological LLM that jointly integrates empathy, expertise, and reasoning. 
To support model development, we design a multi-stage data synthesis pipeline that generates high-quality psychological reasoning samples with detailed rationales and empathetic dialogues.
The reasoning rationales are further enhanced through iterative prompt–rationale optimization, and a multi-LLM cross-selection strategy is employed to identify challenging examples.
Finally, the challenging subset is used for GRPO, while the remaining data are employed for SFT, together contributing to the final model.
Extensive experiments demonstrate that \textit{Psyche-R1} outperforms existing psychological LLMs, achieving performance comparable to \texttt{DeepSeek-R1}.
Moreover, we perform comprehensive ablation studies and analyses to evaluate the individual contributions of each component and strategy within the proposed framework.

\section*{Acknowledgments}

This work was supported in part by National Natural Science Foundation of China under Grant 62402158, Grant 62502145 and Grant 62272144;
by the Key Science \& Technology Project of Anhui Province (202523j08050001);
by the National College Students' Innovation and Entrepreneurship Training Program (202510359110);
by the Anhui Provincial Natural Science Foundation Grant 2408085QF188, and Grant 2408085J040; 
by the Fundamental Research Funds for the Central Universities Grant JZ2025HGTA0162, and Grant JZ2025HGQA0134;
and by the Major Project of Anhui Provincial Science and Technology Breakthrough Program (202423k09020001).

\section*{Limitations}
Despite the promising results of our \textit{Psyche-R1}, our study is subject to several limitations that remain to be addressed in future research.

\boldpara{Language and Cultural Specificity.}
To mitigate the shortage of mental health professionals in China, current \textit{Psyche-R1} and its training corpus are predominantly tailored to the Chinese language and cultural context.
Consequently, the model's empathetic reasoning involves specific cultural norms that may not directly transfer to other languages \cite{adilazuarda-etal-2024-towards, dai2026tearscheersbenchmarkingllms}.
We frame this as a necessary step for local applicability, noting that cross-cultural generalization remains challenging for future research.
Moreover, the integration of multimodal psychology and affective computing warrants further study \cite{song2024emotional,LIAO2026113366}.

\boldpara{Model Scale.}
Constrained by computational resources, \textit{Psyche-R1} is built upon a 7B-parameter backbone.
While it achieves competitive performance, we posit that employing a base model with a larger scale would yield superior performance.
In future work, more efficient fine-tuning strategies \cite{hu2022lora, zhang2025hyperadaloraacceleratinglorarank} may help improve training efficiency and facilitate scaling to stronger backbones under limited resources.

\section*{Ethical Considerations}
The development and deployment of LLMs in the mental health domain necessitate rigorous adherence to ethical standards. 
\boldpara{Nature of the System.}
\textit{Psyche-R1} is designed as a supportive tool for mental health support and education, rather than a replacement for qualified mental health professionals. 
The model is not authorized to provide medical diagnoses, prescribe treatments, or handle crisis interventions. 
Users facing severe mental health crises should seek help from human professionals or emergency services.

\boldpara{Data Privacy and Safety.} 
We prioritize the privacy and safety of individuals in our data curation process. 
For data collected from social media platforms (Type \uppercase\expandafter{\romannumeral4}), we implemented strict de-identification procedures to remove all personally identifiable information, including names, locations, and contact details.
We strictly adhere to data usage policies and ensure that the synthesized data does not reconstruct real-world private interactions.
Furthermore, our data synthesis pipeline that prioritizes high-quality, constructive psychological advice filtered out toxic content and harmful suggestions to align with safety guidelines.

\bibliography{custom}

\begin{thebibliography}{70}
\providecommand{\natexlab}[1]{#1}

\bibitem[{Adilazuarda et~al.(2024)Adilazuarda, Mukherjee, Lavania, Singh, Aji, O{'}Neill, Modi, and Choudhury}]{adilazuarda-etal-2024-towards}
Muhammad~Farid Adilazuarda, Sagnik Mukherjee, Pradhyumna Lavania, Siddhant~Shivdutt Singh, Alham~Fikri Aji, Jacki O{'}Neill, Ashutosh Modi, and Monojit Choudhury. 2024.
\newblock \href {https://doi.org/10.18653/v1/2024.emnlp-main.882} {Towards measuring and modeling ``culture'' in {LLM}s: A survey}.
\newblock In \emph{Proceedings of the 2024 Conference on Empirical Methods in Natural Language Processing}, pages 15763--15784, Miami, Florida, USA. Association for Computational Linguistics.

\bibitem[{Asad et~al.(2019)Asad, Mahmud~Pranto, Afreen, and Islam}]{al2019depression}
Nafiz~Al Asad, Md.~Appel Mahmud~Pranto, Sadia Afreen, and Md.~Maynul Islam. 2019.
\newblock \href {https://doi.org/10.1109/SPICSCON48833.2019.9065101} {Depression detection by analyzing social media posts of user}.
\newblock In \emph{2019 IEEE International Conference on Signal Processing, Information, Communication \& Systems (SPICSCON)}, pages 13--17.

\bibitem[{Chen et~al.(2024)Chen, Cai, Ji, Wang, Liu, Wang, Hou, and Wang}]{chen2024huatuogpt}
Junying Chen, Zhenyang Cai, Ke~Ji, Xidong Wang, Wanlong Liu, Rongsheng Wang, Jianye Hou, and Benyou Wang. 2024.
\newblock \href {https://arxiv.org/abs/2412.18925} {Huatuogpt-o1, towards medical complex reasoning with llms}.
\newblock \emph{Preprint}, arXiv:2412.18925.

\bibitem[{Chen et~al.(2023{\natexlab{a}})Chen, Xing, Lin, Zheng, Wang, Liu, and Xu}]{chen-etal-2023-soulchat}
Yirong Chen, Xiaofen Xing, Jingkai Lin, Huimin Zheng, Zhenyu Wang, Qi~Liu, and Xiangmin Xu. 2023{\natexlab{a}}.
\newblock \href {https://doi.org/10.18653/v1/2023.findings-emnlp.83} {{S}oul{C}hat: Improving {LLM}s' empathy, listening, and comfort abilities through fine-tuning with multi-turn empathy conversations}.
\newblock In \emph{Findings of the Association for Computational Linguistics: EMNLP 2023}, pages 1170--1183, Singapore. Association for Computational Linguistics.

\bibitem[{Chen et~al.(2023{\natexlab{b}})Chen, Lu, and Wang}]{chen-etal-2023-empowering}
Zhiyu Chen, Yujie Lu, and William Wang. 2023{\natexlab{b}}.
\newblock \href {https://doi.org/10.18653/v1/2023.findings-emnlp.284} {Empowering psychotherapy with large language models: Cognitive distortion detection through diagnosis of thought prompting}.
\newblock In \emph{Findings of the Association for Computational Linguistics: EMNLP 2023}, pages 4295--4304, Singapore. Association for Computational Linguistics.

\bibitem[{Cheng et~al.(2024)Cheng, Meng, and Zhang}]{10960233}
Yihang Cheng, Hao Meng, and Wei Zhang. 2024.
\newblock \href {https://doi.org/10.1109/IEIR62538.2024.10960233} {Collaborative agents for anxious depression diagnosis and intervention leverage neuroimages in human brain-ai interactions}.
\newblock In \emph{2024 International Conference on Intelligent Education and Intelligent Research (IEIR)}, pages 1--5.

\bibitem[{Cho et~al.(2023)Cho, Kim, Kim, Kwon, Kwon, Lee, and Lim}]{cho2023evaluating}
Yujin Cho, Mingeon Kim, Seojin Kim, Oyun Kwon, Ryan~Donghan Kwon, Yoonha Lee, and Dohyun Lim. 2023.
\newblock \href {https://arxiv.org/abs/2311.09243} {Evaluating the efficacy of interactive language therapy based on llm for high-functioning autistic adolescent psychological counseling}.
\newblock \emph{Preprint}, arXiv:2311.09243.

\bibitem[{Comanici et~al.(2025)Comanici, Bieber, Schaekermann, Pasupat, Sachdeva, Dhillon, Blistein, Ram, Zhang, Rosen, Marris, Petulla, Gaffney, Aharoni, Lintz, Pais, Jacobsson, Szpektor, Jiang, Haridasan, Omran, Saunshi, Bahri, Mishra, Chu, Boyd, Hekman, Parisi, Zhang, Kawintiranon, Bedrax-Weiss, Wang, Xu, Purkiss, Mendlovic, Deutel, Nguyen, Langley, Korn, Rossazza, Ramé, Waghmare, Miller, Byrd, Sheshan, Hadsell, Bhardwaj, Janus, Rissa, Horgan, Abdagic, Belenki, Allingham, Singh, Guidroz, Srinivasan, Schmit, Chiafullo, Elisseeff, Jha, Kolhar, Berrada, Ding, Si, Mallick, Och, Erell, Ni, Latkar, Yang, Sirkovic, Feng, Leland, Hornung, Wu, Blundell, Alvari, Huang, Yip, Deur, Liu, Surita, Duque, Damen, Jia, Guez, Mircea, Sinha, Magni, Stradomski, Marian, Galić, Chen, Husain, Singhal, Grewe, Aubet, Song, Blanco, Rechis, Ho, Munoz, Zheng, Hamrick, Mather, Taitelbaum, Rutherford, Lei, Chen, Shukla, Moreira, Doi, Isik, Shabat, Rogozińska, Kolipaka, Chang, Vušak, Venkatachary, Noghabi, Bharti, Jun, Zaks, Green,
  Challagundla, Wong, Mohammad, Hirsch, Cheng, Naim, Proleev, Vincent, Singh, Krikun, Krishnan, Ghahramani, Atias, Aggarwal, Kirov, Vytiniotis, Koh, Chronopoulou, Dogra, Ion, Tyen, Lee, Weissenberger, Strohman, Balakrishna, Rae, Velic, de~Liedekerke, Elyada, Yuan, Liu, Shani, Kishchenko, Alessio, Li, Song, Kwei, Jankowski, Pappu, Namiki, Ma, Tripuraneni, Cherry, Ikonomidis, Ling, Ji, Westberg, Wright, Yu, Parkinson, Ramaswamy, Connor, Yeganeh, Grover, Kenwright, Litchev, Apps, Tomala, Halim, Castro-Ros, Li, Boral, Sho, Yarom, Malmi, Klinghoffer, Lin, Ansell, S, Zhao, Zuo, Santoro, Cheng, Demmessie, Liu, Brichtova, Culp, Braun, Graur, Ng, Mehta, Phillips, Sundberg, Godbole, Liu, Katariya, Rim, Seyedhosseini, Ammirati, Valfridsson, Malihi, Knight, Toor, Lampe, Ittycheriah, Chiang, Yeung, Fréchette, Rao, Wang, Srivastava, Zhang, Rhodes, Brand, Weesner, Figotin, Gimeno, Fellinger, Marcenac, Leal, Marcus, Cotruta, Cabrera, Luo, Garrette, Axelrod, Baltateanu, Barker, Chen, Toma, Ingram, Riesa, Kulkarni, Zhang,
  Liu, Wang, Polacek, Wu, Hui, Reyes, Su, Barnes, Malhi, Siddiqui, Feng, Damaschin, Pighin, Steiner, Yang, Boppana, Ivanov, Kandoor, Shah, Mujika, Huang, Choquette-Choo, Patel, Yu, Creswell, Jerry, Liu, Barros, Razeghi, Roy, Culliton, Xiong, Pan, Strohmann, Powell, Seal, DeCarlo, Shyam, Katircioglu, Wang, Hardin, Odisho, Broder, Chang, Nair, Shtefan, O'Brien, Agarwal, Potluri, Goyal, Jhindal, Thakur, Stuken, Lyon, Toutanova, Feng, Wu, Horn, Wang, Cullum, Taubman, Shrivastava, Shi, Tomlinson, Patel, Tu, Oflazer, Pongetti, Yang, Taïga, Perot, Pierse, Han, Drori, Iturrate, Chakrabarti, Yeung, Dopson, ting Chen, Kulshreshtha, Guo, Pham, Schuster, Chen, Polozov, Xing, Zhou, Kacham, Kukliansky, Miech, Yaroshenko, Chi, Douglas, Fei, Blondel, Myla, Madmoni, Wu, Keysers, Kjems, Albuquerque, Yu, D'sa, Plantan, Ionescu, Elias, Gupta, Vuyyuru, Alcober, Zhou, Ji, Hartmann, Puttagunta, Song, Amid, Stefanoiu, Lee, Pucciarelli, Wang, Raul, Petrov, Tian, Anklin, Nti, Gomes, Schumacher, Vesom, Panagopoulos, Bousmalis, Andor,
  Jacob, Zhang, Rosgen, Kecman, Tung, Belias, Goodman, Covington, Wieder, Saxena, Davoodi, Huang, Maddineni, Roulet, Campbell-Ajala, Sessa, Xintian, Wu, Lai, Collins, Haig, Sakenas, Xu, Giustina, Shafey, Charoenpanit, Garg, Ainslie, Severson, Arenas, Pathak, Rajayogam, Feng, Bakker, Li, Wichers, Rogers, Geng, Li, Jagerman, Jia, Olmert, Sharon, Mauger, Mariserla, Ma, Mohabey, Kim, Andreev, Pollom, Love, Jain, Agrawal, Schroecker, Fortin, Warmuth, Liu, Leach, Blok, Girirajan, Aharoni, Uria, Sozanschi, Goldberg, Ionita, Ribeiro, Zlocha, Birodkar, Lachgar, Yuan, Choudhury, Ginsberg, Zheng, Dibb, Graves, Lokhande, Rasskin, Muraru, Quick, Tata, Sermanet, Chawla, Karo, Wang, Zhang, Keller, Dragan, Su, Chou, Liu, Tao, Prabhakara, Wilson, Liu, Wang, Evans, Du, Castaño, Prasad, Mahdy, Gerlach, Reid, Kahn, Zait, Pillai, Ulrich, Wang, Wassenberg, Farkash, Yalasangi, Wang, Bauza, Bucher, Liu, Yan, Leung, Sindhwani, Barnes, Singh, Jurin, Chang, Bhumihar, Eiger, Citovsky, Withbroe, Li, Xue, Santo, Stoyanov, Raimond, Zheng,
  Gao, Listík, Kwasiborski, Saputro, Ozturel, Mallya, Majmundar, West, Caron, Wei, Castrejon, Vikram, Ramachandran, Dhawan, Park, Smoot, van~den Driessche, Blau, Malik, Liang, Hirsch, dos Santos, Weinstein, van~den Oord, Lall, FitzGerald, Jiang, Yang, Webster, Elqursh, Pope, Rotival, Raposo, Zhu, Dean, Alabed, Tran, Gupta, Gleicher, Austin, Rosseel, Umekar, Das, Sun, Chen, Misiunas, Zhou, Di, Loo, Newlan, Li, Ramasesh, Xu, Chen, Gandhe, Soricut, Gupta, Hu, El-Sayed, Garcia, Brusilovsky, Chen, Bolt, Huang, Gurney, Zhang, Pritzel, Wilkiewicz, Seybold, Shamanna, Fischer, Dean, Gill, Mcilroy, Bhowmick, Selier, Yang, Cheng, Magay, Tan, Varma, Walder, Kocisky, Nakashima, Natsev, Kwong, Gog, Zhang, Dieleman, Jimma, Ryabtsev, Brahma, Steiner, Du, Žužul, Žanić, Raghavachari, Gierke, Zheng, Petrova, Dauphin, Liu, Kessler, Hand, Duvarney, Kim, Lee, Hussenot, Hui, Smith, Jain, Xia, Tomar, Amiri, Phan, Fuchs, Weyand, Tomasev, Cordell, Liu, Mallinson, Joshi, Crawford, Suggala, Chien, Fernando, Sanchez-Vargas,
  Williams, Crone, Luo, Karpov, Shan, Thurk, Strudel, Voigtlaender, Patil, Dozat, Khodaei, Singla, Ambroszczyk, Wu, Chang, Roark, Hegde, Ding, Filos, Wu, Pinto, Liu, Khanna, Pandey, Mcloughlin, Li, Haves, Zhou, Buchatskaya, Leal, de~Boursac, Akazawa, Anderson, Chen, Somandepalli, Liang, Goenka, Winkler, Grushetsky, Ding, Smith, Ye, Pont-Tuset, Li, Li, Golany, Wegner, Jiang, Barak, Shangguan, Vértes, Wong, Bornschein, Tudor, Bevilacqua, Schaul, Rawat, Zhao, Axiotis, Meng, McLean, Lai, Beattie, Kushman, Liu, Kutzman, Lang, Ye, Netrapalli, Mishra, Khan, Goel, Willoughby, Tian, Zhuang, Chen, Tsai, Kementsietsidis, Khare, Keeling, Xu, Waters, Altché, Popat, Mittal, Saxton, Badawy, Mathieu, Zheng, Zhou, Ranka, Shin, Duan, Salimans, Mihailescu, Shaham, Chang, Assael, Dikkala, Izzard, Cohen-Addad, Graves, Feinberg, Chung, Strouse, Karmon, Sharifzadeh, Ashwood, Pham, Blanton, Vasiloff, Barber, Geller, Zhou, Zubach, Huang, Zhang, Gupta, Young, Proskurnia, Votel, Gabeur, Barcik, Tripathi, Yu, Yan, Changpinyo,
  Pavetić, Coyle, Fujii, Mendez, Zhou, Rajamani, Hechtman, Cao, Juan, Tan, Dalibard, Du, Clay, Yao, Jia, Vijaykumar, Zhou, Bai, Hung, Pecht, Todorov, Khadke, Gupta, Lahoti, Autef, Duddu, Lee-Thorp, Bykovsky, Misiunas, Flennerhag, Thangaraj, McGiffin, Nado, Kunesch, Noever, Hertz, Liang, Stone, Palmer, Daruki, Pramanik, Põder, Kyker, Khan, Sluzhaev, Ritter, Ruderman, Zhou, Nagpal, Vodrahalli, Necula, Barham, Pavlick, Hartford, Shafran, Zhao, Mikuła, Eccles, Shimokawa, Garg, Vilnis, Chen, Shumailov, Lee, Abdelhamed, Xie, Cohen, Hlavnova, Malkin, Sitawarin, Lottes, Coquinot, Yu, Kumar, Zhang, Mahendru, Ahmed, Martens, Chen, Boag, Peng, Devin, Klimovskiy, Phuong, Vainstein, Xie, Ramabhadran, Howard, Yu, Goswami, Cui, Shleifer, Pinto, Yeh, Yang, Javanmardi, Ethier, Lee, Orbay, Kotecha, Bromberg, Shaw, Thornton, Rosenthal, Gu, Thomas, Gemp, Ayyar, Ushio, Selvan, Wee, Liu, Majzoubi, Yu, Abernethy, Liechty, Pan, Nguyen, Qiong, Hu, Perrin, Arora, Pitler, Wang, Shivakumar, Prost, Limonchik, Wang, Gao, Cour, Buch,
  Gui, Ivanova, Neubeck, Chan, Kim, Chen, Goyal, Chung, Liu, Su, Petrushkina, Shen, Joulin, Xu, Lin, Kulizhskaya, Chelba, Vasudevan, Collins, Bashlovkina, Lu, Fritz, Park, Zhou, Su, Tanburn, Sushkov, Rasquinha, Li, Prendki, Li, LV, Sharma, Fitoussi, Huang, Dai, Dao, Burrows, Prior, Qin, Pundak, Sjoesund, Khurshudov, Zhu, Webson, Kemp, Tan, Agrawal, Sargsyan, Cheng, Stephan, Kwiatkowski, Reid, Byravan, Michaely, Heess, Zhou, Goenka, Carpenter, Levskaya, Wang, Roberts, Leblond, Chikkerur, Ginzburg, Chang, Riachi, Chuqiao, Xu, Borsos, Pliskin, Pawar, Lustman, Kirkwood, Anand, Chaudhary, Kalb, Milan, Augenstein, Goldie, Prince, Raman, Sun, Xia, Cohen, Huo, Camp, Ellis, Zilka, Torres, Patel, Arora, Chan, Adler, Ayoub, Liang, Jamil, Jiang, Baumgartner, Sun, Karov, Akulov, Zheng, Cai, Fantacci, Rubin, Acha, Wang, D'Souza, Sathyanarayana, Dai, Rowe, Simanovsky, Goldman, Kuang, Pan, Rosenberg, Rojas-Esponda, Dutta, Zeng, Jurenka, Farquhar, Bansal, Iqbal, Roelofs, Joung, Beak, Ryu, Poplin, Wu, Alayrac, Buthpitiya,
  Ronneberger, Habtegebriel, Li, Cavallaro, Wei, Bensky, Denk, Ganapathy, Stanway, Joshi, Bertolini, Lo, Ma, Charles, Sampemane, Sahni, Chen, Askham, Gaddy, Young, Tan, Eyal, Bražinskas, Zhong, Wu, Epstein, Bailey, Hard, Lee, Goldshtein, Ruiz, Badawi, Lochbrunner, Kearns, Brown, Pardo, Weber, Yang, Jiang, Akin, Fu, Wainwright, Zou, Gaba, Manzagol, Kan, Song, Zainullina, Lin, Ko, Deshmukh, Jindal, Svensson, Tyam, Zhao, Kaeser-Chen, Baird, Moradi, Hall, Guo, Tsang, Liang, Pereira, Ganesh, Korotkov, Adamek, Thiagarajan, Tran, Chen, Tar, Jain, Dasgupta, Bilal, Reitter, Zhao, Vezzani, Gehman, Mehta, Beltrone, Dotiwalla, Guadarrama, Abbas, Karp, Georgiev, Ferng, Brockschmidt, Peng, Hirnschall, Verma, Bi, Xiao, Dabush, Xu, Wallis, Parker, Wang, Xu, Safarli, Tewari, Zhang, Kim, Gesmundo, Thomas, Levi, Chowdhury, Rao, Garst, Conway-Rahman, Ran, McKinney, Xiao, Yu, Agrawal, Stjerngren, Ionescu, Chen, Sharma, Chiu, Liu, Franko, Sanford, Cai, Michel, Ganapathy, Labanowski, Garrett, Vargas, Sun, Gale, Buschmann,
  Desjardins, Ghelani, Jain, Verma, Asawaroengchai, Eisenschlos, Harlalka, Kazawa, Metzler, Howland, Jian, Ades, Shah, Gangwani, Lee, Ring, Hernandez, Reich, Sinha, Sathe, Kovac, Gill, Kannan, D'olimpio, Sevenich, Whang, Kim, Sim, Chen, Zhang, Lall, Matias, Jia, Friesen, Nasso, Thapliyal, Perozzi, Yu, Shekhawat, Huda, Grabowski, Wang, Sreevatsa, Dib, Hassen, Schuh, Milutinovic, Welty, Quinn, Shah, Wang, Barth-Maron, Frye, Axelsson, Zhu, Ma, Giannoumis, Sedghi, Ye, Luan, Aydin, Chandra, Sampathkumar, Huang, Lavrenko, Eleryan, Hong, Hansen, Carthy, Samanta, Ćevid, Wang, Li, Voznesensky, Hoffman, Terzis, Sehwag, Fidel, He, Cai, He, Feng, Nikoltchev, Phatale, Chase, Lawton, Zhang, Ouyang, Tragut, Manshadi, Narayanan, Shen, Gao, Bolukbasi, Roy, Li, Golovin, Panait, Qin, Han, Anthony, Kudugunta, Patraucean, Ray, Chen, Yang, Bhatia, Talluri, Morris, Ražnatović, Brownfield, An, Peng, Kane, Zheng, Duduta, Kessinger, Noraky, Liu, Rong, Veličković, Rush, Goldin, Wei, Garlapati, Pantofaru, Kwon, Ni, Noland, Trapani,
  Beaufays, Roy, Chow, Turker, Cideron, Mei, Clark, Dou, Bošnjak, Leith, Du, Yazdanbakhsh, Nasr, Kwak, Sheth, Kaskasoli, Anand, Lakshminarayanan, Jerome, Bieber, Chu, Senges, Shen, Sridhar, Ndebele, Beyret, Mohamed, Chen, Freitag, Guo, Liu, Roit, Chen, Yan, Stone, Co-Reyes, Cole, Scellato, Azizi, Hashemi, Jin, Iyer, Valentine, György, Ahuja, Diaz, Lee, Clement, Kong, Garmon, Watts, Bhatia, Gupta, Miecnikowski, Vallet, Taly, Loper, Joshi, Atwood, Chick, Collier, Iliopoulos, Trostle, Gunel, Leal-Cavazos, Hrafnkelsson, Guzman, Ju, Forbes, Emond, Chauhan, Caine, Xiao, Zeng, Moufarek, Murphy, Meng, Gupta, Riedel, Das, Lawal, Narayan, Sosea, Swirhun, Friso, Neyshabur, Lu, Girgin, Wunder, Yvinec, Pyne, Carbune, Rijhwani, Guo, Doshi, Briukhov, Bain, Hitron, Wang, Gupta, Chen, Du, Zhang, Shah, Akula, Dylla, Kachra, Kuo, Zou, Wang, Xu, Zhu, Snyder, Menon, Firat, Mordatch, Yuan, Ponomareva, Blevins, Moore, Wang, Chen, Scholz, Dwornik, Lin, Li, Antognini, I, Song, Miller, Kalra, Raveret, Akerlund, Wu, Nystrom, Godbole,
  Liu, DeBalsi, Zhao, Liu, Caciularu, Lax, Khandelwal, Langston, Bailey, Lattanzi, Wang, Kovelamudi, Mondal, Guruganesh, Hua, Roval, Wesołowski, Ingale, Halcrow, Sohn, Angermueller, Raad, Stickgold, Lu, Kosik, Xie, Lillicrap, Huang, Zhang, Paulus, Farabet, Wertheim, Wang, Joshi, ling Ko, Wu, Agrawal, Lin, Sheng, Sung, Breland-King, Butterfield, Gawde, Singh, Zhang, Apte, Shetty, Hutter, Li, Salesky, Lebron, Kanerva, Paganini, Nguyen, Vallu, Peter, Velury, Kao, Hoover, Bortsova, Bishop, Jakobovits, Agostini, Agarwal, Liu, Kwong, Tavakkol, Bica, Greve, GP, Marcus, Hou, Duerig, Moroshko, Lacey, Davis, Amelot, Wang, Kim, Strinopoulos, Wan, Lan, Krishnan, Tang, Humphreys, Bai, Shtacher, Machado, Pang, Burke, Liu, Aravamudhan, Song, Hirst, Singh, Jou, Bai, Piccinno, Fu, Alazard, Meiri, Winter, Chen, Zhang, Heitkaemper, Lambert, Lee, Frömmgen, Rogulenko, Nair, Niemczyk, Bulyenov, Xu, Shemtov, Zadimoghaddam, Toropov, Wirth, Dai, Gollapudi, Zheng, Kurakin, Lee, Bullard, Serrano, Balazevic, Li, Schalkwyk, Murphy,
  Zhang, Sequeira, Datta, Agrawal, Sutton, Attaluri, Chiang, Farhan, Thornton, Lin, Choma, Nguyen, Dasgupta, Robinson, Comşa, Riley, Pillai, Mustafa, Golan, Zandieh, Lespiau, Porter, Ross, Rajayogam, Agarwal, Venugopalan, Shahriari, Yan, Xu, Tobin, Dubov, Shi, Recasens, Kovsharov, Borgeaud, Dery, Vasanth, Gribovskaya, Qiu, Mahdieh, Skut, Nielsen, Zheng, Yu, Bostock, Gupta, Archer, Rawles, Davies, Svyatkovskiy, Tsai, Halpern, Reisswig, Wydrowski, Chang, Puigcerver, Taege, Li, Schnider, Li, Dena, Xu, Telang, Shi, Zen, Kastner, Ko, Subramaniam, Kumar, Blois, Dai, Wieting, Lu, Zeldes, Xie, Hauth, Ţifrea, Li, El-Husseini, Abolafia, Zhou, Ding, Ghalebikesabi, Guía, Maksai, Ágoston Weisz, Arik, Sukhanov, Świetlik, Jia, Yu, Wang, Brand, Bloxwich, Kirmani, Chen, Go, Sprechmann, Kannen, Carin, Sandhu, Edkins, Nooteboom, Gupta, Maggiore, Azizi, Pritch, Yin, Gupta, Tarlow, Smith, Ivanov, Babaeizadeh, Goel, Kambala, Chu, Kastelic, Liu, Soltau, Stone, Agrawal, Kim, Soparkar, Tadepalli, Bunyan, Soh, Kannan, Kim, Chen,
  Halumi, Roy, Wang, Sercinoglu, Gibson, Bhatnagar, Sano, von Dincklage, Ren, Mitrevski, Olšák, She, Doersch, Jilei, Wang, Liu, Tan, Yakar, Warkentin, Ramirez, Lebsack, Dillon, Mathews, Cobley, Wu, Chen, Simon, Nath, Sainath, Bendebury, Julian, Mankalale, Ćurko, Zacchello, Brown, Sodhia, Howard, Caelles, Gupta, Evans, Bulanova, Katzen, Goldenberg, Tsitsulin, Stanton, Schillings, Kovalev, Fry, Shah, Lin, Upadhyay, Li, Radpour, Maggioni, Xiong, Haas, Brennan, Kamath, Savinov, Nagrani, Yacovone, Kappedal, Andriopoulos, Lao, Li, Rozhdestvenskiy, Hashimoto, Audibert, Austin, Rodriguez, Ruoss, Honke, Karkhanis, Xiong, Wei, Huang, Leng, Premachandran, Bileschi, Evangelopoulos, Mensink, Pavagadhi, Teplyashin, Chang, Xue, Tanzer, Goldman, Patel, Li, Wiesner, Zheng, Stewart-Binks, Han, Li, Luo, Lenc, Lučić, Xue, Mullins, Guseynov, Chang, Galatzer-Levy, Zhang, Bingham, Hu, Hartman, Ma, Griffith, Irpan, Radebaugh, Yue, Fan, Ungureanu, Sorokin, Teufel, Li, Anil, Paparas, Wang, Lin, Peng, Shum, Petrovic, Brady,
  Nguyen, Macherey, Li, Singh, Yenugula, Iinuma, Chen, Kopparapu, Stern, Dave, Thekkath, Perot, Kumar, Li, Xiao, Bilotti, Bateni, Noble, Lee, Vázquez-Reina, Salazar, Yang, Wang, Gruzewska, Rao, Raghuram, Xu, Ben-David, Mei, Dalmia, Zhang, Liu, Bansal, Pankov, Schwarcz, Burns, Chan, Sanghai, Liang, Liang, He, Stuart, Narayanan, Zhu, Frank, Fatemi, Sabne, Lang, Bhattacharya, Settle, Wang, McMahan, Tacchetti, Soares, Hadian, Cabi, Chung, Putikhin, Li, Chen, Tarango, Michalewski, Kazemi, Masoom, Sheftel, Shivanna, Vadali, Comanescu, Reid, Moore, Neelakantan, Sander, Herzig, Rosenberg, Dehghani, Choi, Fink, Hayes, Ge, Weng, Ho, Karro, Krishna, Thiet, Skerry-Ryan, Eppens, Andreetto, Sarma, Bonacina, Ayan, Nawhal, Shan, Dusenberry, Thakoor, Gubbi, Nguyen, Tsarfaty, Albanie, Mitrović, Gandhi, Chen, Epasto, Stephanov, Jin, Gehman, Amini, Weber, Behbahani, Xu, Allamanis, Chen, Ott, Sha, Jastrzebski, Qi, Greene, Wu, Toki, Vlasic, Shapiro, Kotikalapudi, Shen, Saeki, Xie, Cassirer, Bharadwaj, Kiyono, Bhojanapalli,
  Rosenfeld, Ritter, Mao, Oliveira, Egyed, Bandemer, Parisotto, Kinoshita, Pluto, Maniatis, Li, Guo, Ghiasi, Tarbouriech, Chatterjee, Jin, Katrina, Xu, Palomaki, Arnold, Sewak, Piccinini, Sharma, Albrecht, Purser-haskell, Vaswani, Chen, Wisniewski, Cao, Aslanides, Phu, Sieb, Agubuzu, Zheng, Sohn, Selvi, Andreassen, Subudhi, Eruvbetine, Woodman, Mery, Krause, Ren, Ma, Luo, Chen, Fan, Griffiths, Schuler, Li, Zhang, Sarr, Luo, Patana, Watson, Naboulsi, Collins, Sidhwani, Hoogeboom, Silver, Caveness, Zhao, Rodriguez, Deines, Bai, Griffin, Tagliasacchi, Xue, Babbula, Pang, Ding, Shen, Peake, Crocker, Raghvendra, Swisher, Han, Singh, Wu, Pchelin, Munkhdalai, Alon, Bacon, Robles, Bulian, Johnson, Powell, Ferreira, Li, Benzing, Velimirović, Soyer, Kong, Tony, Nguyên, Yang, Liu, van Amersfoort, Gillick, Sun, Rauschmayr, Zhang, Zhan, Zhou, Frolov, Yang, Vnukov, Rouillard, Li, Mandhane, Fallen, Venkataraman, Hu, Brennan, Lee, Chang, Sundermeyer, Pan, Ke, Tong, Fabrikant, Bono, Gu, Foley, Mao, Delakis, Bhaswar,
  Frostig, Li, Zipori, Hope, Kozlova, Mishra, Djolonga, Schiff, Merey, Briakou, Morgan, Wan, Hassidim, Skerry-Ryan, Sengupta, Jasarevic, Kallakuri, Kunkle, Brennan, Lieber, Mansoor, Walker, Zhang, Xie, Žužić, Chukwuka, Druinsky, Cho, Yao, Naeem, Butt, Kim, Jia, Jordan, Lelkes, Kurzeja, Wang, Zhao, Over, Chakladar, Prasetya, Jha, Ganapathy, Cong, Shroff, Saroufim, Miryoosefi, Hammad, Nasir, Xi, Gao, Maeng, Hora, Cheng, Haghani, Lewenberg, Lu, Matysiak, Raisinghani, Wang, Baugher, Sukthankar, Giang, Schultz, Fiedel, Chen, Lee, Dey, Zheng, Paul, Smith, Ly, Wang, Bansal, Perz, Ricco, Blank, Keshava, Sharma, Chow, Lad, Jalan, Osindero, Swanson, Scott, Ilić, Li, Jonnalagadda, Soudagar, Xiong, Batsaikhan, Jarrett, Kumar, Shah, Lawlor, Waters, Graham, May, Ramos, Lefdal, Cankara, Cano, O'Donoghue, Borovik, Liu, Grimstad, Alnahlawi, Tsihlas, Hudson, Grigorev, Jia, Huang, Igwe, Lebedev, Tang, Krivokon, Garcia, Tan, Jia, Stys, Vashishth, Liang, Venkatraman, Gu, Kementsietsidis, Zhu, Jung, Bai, Hosseini, Ahmed,
  Gupta, Yuan, Ashraf, Nigam, Vasudevan, Awasthi, Gilady, Mariet, Eskander, Li, Hu, Garrido, Schlattner, Zhang, Saxena, Dević, Muralidharan, Murthy, Zhou, Choi, Wongpanich, Wang, Shah, Xu, Huang, Spencer, Chen, Cohan, Wang, Tompson, Wu, Haroun, Li, Huergo, Yang, Yin, Wendt, Bendersky, Chaabouni, Snaider, Ferret, Jindal, Thompson, Xue, Bishop, Phal, Sharma, Sung, Radhakrishnan, Shomrat, Ingle, Vij, Gilmer, Istin, Sobell, Lu, Nottage, Sadigh, Willcock, Zhang, Xu, Brown, Lee, Wang, Zhu, Tay, Kim, Gutierrez, Sharma, Xian, Seo, Cui, Pochernina, Baetu, Jastrzębski, Ly, Elhawaty, Suh, Sezener, Wang, Yuen, Tucker, Cai, Yang, Wang, Muzio, Qian, Yoo, Lockhart, McKee, Guo, Mehrotra, Mendonça, Mehta, Ben, Tekur, Mu, Zhu, Krakovna, Lee, Maschinot, Cevey, Choe, Bai, Srinivasan, Gasaway, Young, Siegler, Holtmann-Rice, Piratla, Baumli, Yogev, Hofer, van Hasselt, Grant, Chervonyi, Silver, Hogue, Agarwal, Wang, Singh, Flynn, Lipschultz, David, Bellot, Yang, Le, Graziano, Olszewska, Hui, Maurya, Parotsidis, Chen, Oguntebi,
  Kelley, Baddepudi, Mauerer, Shaw, Siegman, Yang, Shetty, Roy, Song, Stokowiec, Burnell, Savant, Busa-Fekete, Miao, Ghosh, MacDermed, Lippe, Dektiarev, Behrman, Mentzer, Nguyen, Wei, Verma, Knutsen, Dasari, Yan, Mitrichev, Wang, Shejwalkar, Austin, Sunkara, Potti, Virin, Wright, Liu, Riva, Pot, Kochanski, Le, Balasubramaniam, Dhar, Liao, Bloniarz, Shukla, Cole, Lee, Zhang, Kafle, Vashishtha, Mahmoudieh, Chen, Hoffmann, Srinivasan, Lago, Shalom, Wang, Elabd, Sharma, Oh, Kothawade, Le, Monteiro, Yang, Alarakyia, Geirhos, Mincu, Garnes, Kobayashi, Mariooryad, Krasowiak, Zhixin, Lai, Mourad, Wang, Bu, Aharoni, Chen, Goyal, Zubov, Bapna, Dabir, Kothari, Lamerigts, Cao, Shar, Yew, Kulkarni, Mahaarachchi, Joshi, Zhu, Lichtarge, Zhou, Muckenhirn, Selo, Vinyals, Chen, Brohan, Mehta, Cogan, Wang, Geri, Ko, Chen, Viola, Shivam, Wang, Elish, Popa, Pereira, Liu, Koster, Kim, Zhang, Ebrahimi, Talukdar, Zheng, Poklukar, Mikhalap, Johnson, Vijayakumar, Omernick, Dibb, Dubey, Hu, Suman, Aggarwal, Kornakov, Xia, Lowe,
  Kolganov, Xiao, Nikolaev, Hemingray, Li, Iljazi, Rybiński, Sandhu, Lu, Luong, Jenatton, Govindaraj, Hui, Li, Dulac-Arnold, Park, Wang, Modi, Pouget-Abadie, Greller, Gupta, Berry, Ramachandran, Xie, McCafferty, Wang, Gupta, Lim, Bratanič, Brock, Akolzin, Sproch, Karliner, Kim, Goedeckemeyer, Shazeer, Schmid, Calandriello, Bhatia, Choromanski, Montgomery, Dua, Ramalho, King, Gao, Nguyen, Lindner, Pitta, Johnson, Salama, Ardila, Han, Farnese, Odoom, Wang, Ding, Rink, Smith, Lehri, Cohen, Vats, He, Gopavarapu, Paszke, Patel, Gansbeke, Loher, Castro, Voitovich, von Glehn, George, Niklaus, Eaton-Rosen, Rakićević, Jue, Perel, Zhang, Bahat, Pouget, Xing, Huot, Shenoy, Bos, Coriou, Richter, Noy, Wang, Ontanon, Qin, Makarchuk, Hassabis, Li, Sharma, Venkatesan, Kemaev, Daniel, Huang, Shah, Ponce, Warren, Chen, Faruqui, Wu, Andačić, Payrits, McDuff, Hume, Cao, Tessler, Wang, Wang, Rendulic, Agustsson, Johnson, Lando, Howard, Padmanabhan, Daswani, Banino, Kilgore, Heek, Ji, Caceres, Li, Kassner, Vlaskin, Liu,
  Grills, Hou, Sukkerd, Cheon, Shetty, Markeeva, Stanczyk, Iyer, Gong, Gao, Gopalakrishnan, Blyth, Reynolds, Bhoopchand, Bilenko, Gharibian, Zayats, Faust, Singh, Ma, Jiao, Vijayanarasimhan, Aroyo, Yadav, Chakera, Kakarla, Meshram, Gregor, Botea, Senter, Jia, Kovacs, Sharma, Baur, Kang, He, Zhuo, Kostelac, Laish, Peng, O'Bryan, Kasenberg, Rao, Leurent, Zhang, Stevens, Salazar, Zhang, Lobov, Walker, Porter, Redshaw, Ke, Rao, Lee, Lam, Moffitt, Kim, Qiao, Koo, Dadashi, Song, Sundararajan, Xu, Kawamoto, Zhong, Barbu, Reddy, Verzetti, Li, Papamakarios, Klimczak-Plucińska, Cassin, Kavukcuoglu, Swavely, Vaucher, Zhao, Hemsley, Tschannen, Ge, Menghani, Yu, Ha, He, Wu, Song, Sterneck, Zinke, Calian, Marsden, Ruiz, Hessel, Gueta, Lee, Farris, Gupta, Li, Saleh, Misra, Xiao, Mendolicchio, Buttimore, Krayvanova, Nayakanti, Wiethoff, Pande, Mirhoseini, Lao, Liu, Hua, Chen, Malkov, Kalashnikov, Gupta, Audhkhasi, Zhai, Kopalle, Jain, Ofek, Meyer, Baatarsukh, Strejček, Qian, Freedman, Figueira, Sokolik, Bachem, Lin,
  Kharrat, Hidey, Xu, Duan, Li, Ersoy, Everett, Cen, Santamaria-Fernandez, Taubenfeld, Mackinnon, Deng, Zablotskaia, Viswanadha, Goel, Yates, Deng, Choy, Chen, Sinha, Mossin, Wang, Szlam, Hao, Rubenstein, Toksoz-Exley, Aperghis, Zhong, Ahn, Isard, Lacombe, Luisier, Anastasiou, Kalley, Prabhu, Dunleavy, Bijwadia, Mao-Jones, Chen, Pasumarthi, Wood, Dostmohamed, Hurley, Simsa, Parrish, Pajarskas, Harvey, Skopek, Kochinski, Rey, Rieser, Zhou, Lee, Acharya, Li, Jiang, Zhang, Gipson, Mahintorabi, Gelmi, Khajehnouri, Yeh, Lee, Matthey, Baker, Pham, Fu, Pak, Gupta, Vasconcelos, Sadovsky, Walker, Hsiao, Zochbauer, Marzoca, Velan, Zeng, Baechler, Driess, Jain, Huang, Tao, Maggs, Levine, Schneider, Gemzer, Petit, Han, Fisher, Zelle, Biles, Ie, Fadeeva, Liu, Franco, Collister, Zhang, Wang, Zhao, Kieliger, Shuster, Zhu, Gong, Chan, Sun, Basu, Zimmermann, Hayes, Bapna, Snoek, Yang, Datta, Abdallah, Kilgour, Li, Mah, Jun, Rivière, Karmarkar, Spalink, Huang, Gonzalez, Tran, Nowak, Palowitch, Chadwick, Talius, Mehta, Sellam,
  Fränken, Nicosia, He, Kini, Amos, Basu, Jobe, Shaw, Xu, Evans, Ikeda, Yan, Jin, Wang, Yadav, Labzovsky, Sampath, Ma, Schumann, Siddhant, Shah, Youssef, Agarwal, Dabney, Tonioni, Ambar, Li, Guyon, Li, Soergel, Fang, Karadzhov, Udrescu, Trinh, Raunak, Noury, Guo, Gupta, Finkelstein, Petek, Liang, Billock, Sun, Wood, Song, Yu, Matejovicova, Cohen, Andra, D'Ambrosio, Deng, Nallatamby, Songhori, Dangovski, Lampinen, Botadra, Hillier, Cao, Baddi, Kuncoro, Yoshino, Bhagatwala, Ranzato, Schaeffer, Liu, Ye, Sarvana, Nham, Kuang, Gao, Baek, Mittal, Wahid, Gergely, Ni, Feldman, Muir, Lamblin, Macherey, Dyer, Kilpatrick, Campos, Bhutani, Fort, Ahmad, Severyn, Chatziprimou, Ferludin, Dimarco, Kusupati, Heyward, Bahir, Villela, Millican, Marcus, Bahargam, Unlu, Roth, Wei, Gopal, Ghoshal, Lee, Lin, Lees, Lee, Hosseini, Fan, Neel, Wu, Altun, Cai, Piqueras, Woodward, Bissacco, Haykal, Bordbar, Sundaram, Hodkinson, Toyama, Polovets, Myers, Sinha, Levinboim, Krishnakumar, Chhaparia, Sholokhova, Gundavarapu, Jawahar, Qureshi,
  Hu, Momchev, Rahtz, Wu, S, Dhamdhere, Guo, Gupta, Eslami, Schain, Blokzijl, Welling, Orr, Bolelli, Perez-Nieves, Sirotenko, Prasad, Kar, Pigem, Terzi, Weisz, Ghosh, Mavalankar, Madeka, Daugaard, Adam, Shah, Berman, Tran, Baker, Andrejczuk, Chole, Raboshchuk, Mirzazadeh, Kagohara, Wu, Schallhart, Orlando, Wang, Rrustemi, Xiong, Liu, Vezer, Ramsden, yiin Chang, Mudgal, Li, Vieillard, Hoshen, Ahmad, Slone, Hua, Potikha, Rossini, Stritar, Prakash, Wang, Dong, Nazari, Nehoran, Tekelioglu, Li, Badola, Funkhouser, Li, Yerram, Ganeshan, Formoso, Langner, Shi, Li, Yamamori, Panda, Saade, Scarpati, Breaux, Carey, Zhou, Hsieh, Bridgers, Butryna, Gupta, Tulsyan, Woo, Eltyshev, Grathwohl, Parks, Benjamin, Panigrahy, Dodhia, Freitas, Sauer, Song, Alet, Tolins, Paduraru, Zhou, Albert, Zhang, Shu, Bansal, Nguyen, Globerson, Xiao, Manyika, Hennigan, Rong, Matak, Bakalov, Sharma, Sinopalnikov, Pierson, Roller, Brown, Gao, Fukuzawa, Ghafouri, Vassigh, Barr, Wang, Korsun, Jayaram, Ren, Zaman, Khan, Lunts, Deutsch, Uthus, Katz,
  Samsikova, Khalifa, Sethi, Sun, Tang, Alon, Luo, Yu, Nayyar, Petrini, Truong, Hellendoorn, Chinaev, Alberti, Wang, Hu, Mirrokni, Balashankar, Aharon, Mehta, Iscen, Kready, Manning, Mohananey, Chen, Tripathi, Wu, Petrovski, Hwang, Baeuml, Chandrakaladharan, Liu, Coaguila, Chen, Ma, Tafti, Tatineni, Spitz, Ye, Vicol, Rosca, Puigdomènech, Yahav, Ghemawat, Lin, Kirk, Nabulsi, Brin, Bohnet, Caluwaerts, Veerubhotla, Zheng, Dai, Petrov, Xu, Mehran, Xu, Zintgraf, Choi, Hombaiah, Thoppilan, Reddi, Lew, Li, Webster, Sawhney, Lamprou, Shakeri, Lunayach, Chen, Bagri, Salcianu, Chen, Donchev, Magister, Nørly, Rodrigues, Izo, Noga, Zou, Köppe, Zhou, Lee, Long, Eisenbud, Chen, Schenck, To, Zhong, Taropa, Truong, Levy, Martins, Zhang, Semturs, Zhang, Yakubovich, Moreno, McConnaughey, Lu, Redmond, Weerts, Bitton, Refice, Lacasse, Conmy, Tallec, Odell, Forbes-Pollard, Socala, Hoech, Kohli, Walton, Wang, Sazanovich, Zhu, Kapishnikov, Galt, Denton, Murdoch, Sikora, Mohamed, Wei, First, McConnell, Cobo, Qin, Avrahami, Balle,
  Watanabe, Louis, Kraft, Ariafar, Gu, Rives, Yoon, Rusu, Cobon-Kerr, Hahn, Luo, Yuvein, Zhu, Ahuja, Benenson, Kaufman, Yu, Hightower, Zhang, Ni, Hendricks, Wang, Yona, Jain, Barrio, Bhupatiraju, Velusamy, Dafoe, Riedel, Thomas, Yuan, Bellaiche, Panthaplackel, Kloboves, Jauhari, Akbulut, Davchev, Gladchenko, Madras, Chuklin, Hill, Yuan, Madhavan, Leonhard, Scandinaro, Chen, Niu, Douillard, Damoc, Onoe, Pedregosa, Bertsch, Leichner, Pagadora, Malmaud, Ponda, Twigg, Duzhyi, Shen, Wang, Garg, Chen, Evci, Lee, Liu, Kojima, Yamaguchi, Rajendran, Piergiovanni, Rajendran, Fornoni, Ibagon, Ragan, Khan, Blitzer, Bunner, Sun, Kosakai, Lundberg, Elue, Guu, Park, Park, Narayanaswamy, Wu, Mudigonda, Cohn, Mu, Kumar, Graesser, Zhang, Killam, Zhuang, Giménez, Jishi, Ley-Wild, Zhai, Osawa, Cedillo, Liu, Upadhyay, Sieniek, Sharma, Paine, Angelova, Addepalli, Parada, Majumder, Lamp, Kumar, Deng, Myaskovsky, Sabolić, Dudek, York, de~Chaumont~Quitry, Nie, Cattle, Gunjan, Piot, Khawaja, Bang, Wang, Khodadadeh, R, Rawlani,
  Powell, Lee, Griesser, Oh, Magalhaes, Li, Tokumine, Vogel, Hsu, BC, Jindal, Cohen, Yang, Yuan, de~Cesare, Bruguier, Xu, Roy, Jacovi, Belov, Arya, Meadowlark, Cohen-Ganor, Ye, Morris-Suzuki, Banzal, Song, Ponnuramu, Zhang, Scrivener, Zaiem, Rochman, Han, Ghazi, Lee, Drath, Suo, Girgis, Shenoy, Nguyen, Eck, Gupta, Yan, Carreira, Gulati, Sang, Mirylenka, Cooney, Chou, Ling, Fan, Coleman, Tubone, Kumar, Baldridge, Hernandez-Campos, Lazaridou, Besley, Yona, Bulut, Wellens, Pierigiovanni, George, Green, Han, Tao, Clark, You, Abdolmaleki, Fu, Chen, Chaugule, Chandorkar, Rahman, Thompson, Koanantakool, Bernico, Ren, Vlasov, Vassilvitskii, Kula, Liang, Kim, Huang, Ye, Lepikhin, and Helmholz}]{comanici2025gemini}
Gheorghe Comanici, Eric Bieber, Mike Schaekermann, Ice Pasupat, Noveen Sachdeva, Inderjit Dhillon, Marcel Blistein, Ori Ram, Dan Zhang, Evan Rosen, Luke Marris, Sam Petulla, Colin Gaffney, Asaf Aharoni, Nathan Lintz, Tiago~Cardal Pais, Henrik Jacobsson, Idan Szpektor, Nan-Jiang Jiang, and 3416 others. 2025.
\newblock \href {https://arxiv.org/abs/2507.06261} {Gemini 2.5: Pushing the frontier with advanced reasoning, multimodality, long context, and next generation agentic capabilities}.
\newblock \emph{Preprint}, arXiv:2507.06261.

\bibitem[{Dai et~al.(2026)Dai, Shen, Gao, Li, Jiang, Wang, Liu, Ge, and Hu}]{dai2026tearscheersbenchmarkingllms}
Chongyuan Dai, Yaling Shen, Zihan Gao, Jia Li, Yishun Jiang, Yaxiong Wang, Liu Liu, Zongyuan Ge, and Jinpeng Hu. 2026.
\newblock \href {https://doi.org/10.18653/v1/2026.acl-long.1769} {Tears or cheers? benchmarking {LLM}s via culturally elicited distinct affective responses}.
\newblock In \emph{Proceedings of the 64th Annual Meeting of the {A}ssociation for {C}omputational {L}inguistics (Volume 1: Long Papers)}, pages 38171--38196, San Diego, California, United States. Association for Computational Linguistics.

\bibitem[{Demszky et~al.(2023)Demszky, Yang, Yeager, Bryan, Clapper, Chandhok, Eichstaedt, Hecht, Jamieson, Johnson et~al.}]{demszky2023using}
Dorottya Demszky, Diyi Yang, David~S Yeager, Christopher~J Bryan, Margarett Clapper, Susannah Chandhok, Johannes~C Eichstaedt, Cameron Hecht, Jeremy Jamieson, Meghann Johnson, and 1 others. 2023.
\newblock \href {https://doi.org/10.1038/s44159-023-00241-5} {Using large language models in psychology}.
\newblock \emph{Nature Reviews Psychology}, 2(11):688--701.

\bibitem[{Gao et~al.(2023)Gao, Madaan, Zhou, Alon, Liu, Yang, Callan, and Neubig}]{pmlr-v202-gao23f}
Luyu Gao, Aman Madaan, Shuyan Zhou, Uri Alon, Pengfei Liu, Yiming Yang, Jamie Callan, and Graham Neubig. 2023.
\newblock \href {https://proceedings.mlr.press/v202/gao23f.html} {{PAL}: Program-aided language models}.
\newblock In \emph{Proceedings of the 40th International Conference on Machine Learning}, volume 202 of \emph{Proceedings of Machine Learning Research}, pages 10764--10799. PMLR.

\bibitem[{Guo et~al.(2025)Guo, Yang, Zhang, Song, Wang, Zhu, Xu, Zhang, Ma, Bi, Zhang, Yu, Wu, Wu, Gou, Shao, Li, Gao, Liu, Xue, Wang, Wu, Feng, Lu, Zhao, Deng, Ruan, Dai, Chen, Ji, Li, Lin, Dai, Luo, Hao, Chen, Li, Zhang, Xu, Ding, Gao, Qu, Li, Guo, Li, Chen, Yuan, Tu, Qiu, Li, Cai, Ni, Liang, Chen, Dong, Hu, You, Gao, Guan, Huang, Yu, Wang, Zhang, Zhao, Wang, Zhang, Xu, Xia, Zhang, Zhang, Tang, Zhou, Li, Wang, Li, Tian, Huang, Zhang, Wang, Chen, Du, Ge, Zhang, Pan, Wang, Chen, Jin, Chen, Lu, Zhou, Chen, Ye, Wang, Yu, Zhou, Pan, Li, Zhou, Wu, Yun, Pei, Sun, Wang, Zeng, Liu, Liang, Gao, Yu, Zhang, Xiao, An, Liu, Wang, Chen, Nie, Cheng, Liu, Xie, Liu, Yang, Li, Su, Lin, Li, Jin, Shen, Chen, Sun, Wang, Song, Zhou, Wang, Shan, Li, Wang, Wei, Zhang, Xu, Li, Zhao, Sun, Wang, Yu, Zhang, Shi, Xiong, He, Piao, Wang, Tan, Ma, Liu, Guo, Ou, Wang, Gong, Zou, He, Xiong, Luo, You, Liu, Zhou, Zhu, Huang, Li, Zheng, Zhu, Ma, Tang, Zha, Yan, Ren, Ren, Sha, Fu, Xu, Xie, Zhang, Hao, Ma, Yan, Wu, Gu, Zhu, Liu, Li, Xie, Song,
  Pan, Huang, Xu, Zhang, and Zhang}]{guo2025deepseek}
Daya Guo, Dejian Yang, Haowei Zhang, Junxiao Song, Peiyi Wang, Qihao Zhu, Runxin Xu, Ruoyu Zhang, Shirong Ma, Xiao Bi, Xiaokang Zhang, Xingkai Yu, Yu~Wu, Z.~F. Wu, Zhibin Gou, Zhihong Shao, Zhuoshu Li, Ziyi Gao, Aixin Liu, and 175 others. 2025.
\newblock \href {https://doi.org/10.1038/s41586-025-09422-z} {Deepseek-r1 incentivizes reasoning in llms through reinforcement learning}.
\newblock \emph{Nature}, 645(8081):633–638.

\bibitem[{Hsieh et~al.(2023)Hsieh, Li, Yeh, Nakhost, Fujii, Ratner, Krishna, Lee, and Pfister}]{hsieh-etal-2023-distilling}
Cheng-Yu Hsieh, Chun-Liang Li, Chih-kuan Yeh, Hootan Nakhost, Yasuhisa Fujii, Alex Ratner, Ranjay Krishna, Chen-Yu Lee, and Tomas Pfister. 2023.
\newblock \href {https://doi.org/10.18653/v1/2023.findings-acl.507} {Distilling step-by-step! outperforming larger language models with less training data and smaller model sizes}.
\newblock In \emph{Findings of the Association for Computational Linguistics: ACL 2023}, pages 8003--8017, Toronto, Canada. Association for Computational Linguistics.

\bibitem[{Hu et~al.(2022)Hu, yelong shen, Wallis, Allen-Zhu, Li, Wang, Wang, and Chen}]{hu2022lora}
Edward~J Hu, yelong shen, Phillip Wallis, Zeyuan Allen-Zhu, Yuanzhi Li, Shean Wang, Lu~Wang, and Weizhu Chen. 2022.
\newblock \href {https://openreview.net/forum?id=nZeVKeeFYf9} {Lo{RA}: Low-rank adaptation of large language models}.
\newblock In \emph{International Conference on Learning Representations}.

\bibitem[{Hu et~al.(2025{\natexlab{a}})Hu, Dong, Luo, Ma, Zou, Sun, Guo, Yang, and Wang}]{hu2024psycollm}
Jinpeng Hu, Tengteng Dong, Gang Luo, Hui Ma, Peng Zou, Xiao Sun, Dan Guo, Xun Yang, and Meng Wang. 2025{\natexlab{a}}.
\newblock \href {https://doi.org/10.1109/TCSS.2024.3497725} {Psycollm: Enhancing llm for psychological understanding and evaluation}.
\newblock \emph{IEEE Transactions on Computational Social Systems}, 12(2):539--551.

\bibitem[{Hu et~al.(2025{\natexlab{b}})Hu, Shi, Dai, Li, Song, and Wang}]{hu2025beyond}
Jinpeng Hu, Hongchang Shi, Chongyuan Dai, Zhuo Li, Peipei Song, and Meng Wang. 2025{\natexlab{b}}.
\newblock \href {https://doi.org/10.1145/3746027.3755726} {Beyond emotion recognition: A multi-turn multimodal emotion understanding and reasoning benchmark}.
\newblock In \emph{Proceedings of the 33rd ACM International Conference on Multimedia}, MM '25, page 5814–5823, New York, NY, USA. Association for Computing Machinery.

\bibitem[{Hu et~al.(2026)Hu, Wang, Xie, Li, Ma, and Guo}]{Hu_Wang_Xie_Li_Ma_Guo_2026}
Jinpeng Hu, Ao~Wang, Qianqian Xie, Zhuo Li, Hui Ma, and Dan Guo. 2026.
\newblock \href {https://doi.org/10.1609/aaai.v40i37.40365} {Agentmental: An interactive multi-agent framework for explainable and adaptive mental health assessment}.
\newblock \emph{Proceedings of the AAAI Conference on Artificial Intelligence}, 40(37):31050--31058.

\bibitem[{Huang et~al.(2020)Huang, Epps, Joachim, and Sethu}]{8882493}
Zhaocheng Huang, Julien Epps, Dale Joachim, and Vidhyasaharan Sethu. 2020.
\newblock \href {https://doi.org/10.1109/JSTSP.2019.2949419} {Natural language processing methods for acoustic and landmark event-based features in speech-based depression detection}.
\newblock \emph{IEEE Journal of Selected Topics in Signal Processing}, 14(2):435--448.

\bibitem[{Kuribayashi et~al.(2024)Kuribayashi, Oseki, and Baldwin}]{kuribayashi-etal-2024-psychometric}
Tatsuki Kuribayashi, Yohei Oseki, and Timothy Baldwin. 2024.
\newblock \href {https://doi.org/10.18653/v1/2024.findings-naacl.129} {Psychometric predictive power of large language models}.
\newblock In \emph{Findings of the Association for Computational Linguistics: NAACL 2024}, pages 1983--2005, Mexico City, Mexico. Association for Computational Linguistics.

\bibitem[{Lai et~al.(2024)Lai, Shi, Du, Wu, Fu, Dou, and Wang}]{lai2023supporting}
Tin Lai, Yukun Shi, Zicong Du, Jiajie Wu, Ken Fu, Yichao Dou, and Ziqi Wang. 2024.
\newblock \href {https://doi.org/10.3390/biomedinformatics4010002} {Supporting the demand on mental health services with ai-based conversational large language models (llms)}.
\newblock \emph{BioMedInformatics}, 4(1):8--33.

\bibitem[{Lee et~al.(2020)Lee, Park, Kang, Choi, and Han}]{lee-etal-2020-cross}
Daeun Lee, Soyoung Park, Jiwon Kang, Daejin Choi, and Jinyoung Han. 2020.
\newblock \href {https://doi.org/10.18653/v1/2020.findings-emnlp.200} {Cross-lingual suicidal-oriented word embedding toward suicide prevention}.
\newblock In \emph{Findings of the Association for Computational Linguistics: EMNLP 2020}, pages 2208--2217, Online. Association for Computational Linguistics.

\bibitem[{Lee et~al.(2024)Lee, Kim, Kim, Kang, Yang, Kim, Kang, Jung, Kim, Lee, Chung, Yu, Lee, and Yeo}]{lee-etal-2024-cactus}
Suyeon Lee, Sunghwan Kim, Minju Kim, Dongjin Kang, Dongil Yang, Harim Kim, Minseok Kang, Dayi Jung, Min~Hee Kim, Seungbeen Lee, Kyong-Mee Chung, Youngjae Yu, Dongha Lee, and Jinyoung Yeo. 2024.
\newblock \href {https://doi.org/10.18653/v1/2024.findings-emnlp.832} {Cactus: Towards psychological counseling conversations using cognitive behavioral theory}.
\newblock In \emph{Findings of the Association for Computational Linguistics: EMNLP 2024}, pages 14245--14274, Miami, Florida, USA. Association for Computational Linguistics.

\bibitem[{Liao et~al.(2026)Liao, Zeng, Song, Zhou, Fan, and Wang}]{LIAO2026113366}
Junjie Liao, Jiandian Zeng, Binbin Song, Mengting Zhou, Xiaopeng Fan, and Tian Wang. 2026.
\newblock \href {https://doi.org/10.1016/j.patcog.2026.113366} {Unlocking explainable and effective multimodal affective reasoning via large language models}.
\newblock \emph{Pattern Recognition}, 178:113366.

\bibitem[{Lin(2004)}]{lin-2004-rouge}
Chin-Yew Lin. 2004.
\newblock \href {https://aclanthology.org/W04-1013/} {{ROUGE}: A package for automatic evaluation of summaries}.
\newblock In \emph{Text Summarization Branches Out}, pages 74--81, Barcelona, Spain. Association for Computational Linguistics.

\bibitem[{Liu et~al.(2025)Liu, Wang, Pan, Wan, Dai, Lin, Bai, Rueckert, and Arcucci}]{liu2025beyond}
Che Liu, Haozhe Wang, Jiazhen Pan, Zhongwei Wan, Yong Dai, Fangzhen Lin, Wenjia Bai, Daniel Rueckert, and Rossella Arcucci. 2025.
\newblock \href {https://openreview.net/forum?id=pZ4WkOMDkB} {Beyond distillation: Pushing the limits of medical {LLM} reasoning with minimalist rule-based {RL}}.
\newblock In \emph{The Second Workshop on GenAI for Health: Potential, Trust, and Policy Compliance}.

\bibitem[{Liu et~al.(2026)Liu, Wang, Shi, Xie, An, Zhang, Zhao, Gu, Lin, Hu, Li, Zhang, Zhou, and Gai}]{liu2026attention}
Runze Liu, Jiakang Wang, Yuling Shi, Zhihui Xie, Chenxin An, Kaiyan Zhang, Jian Zhao, Xiaodong Gu, Lei Lin, Wenping Hu, Xiu Li, Fuzheng Zhang, Guorui Zhou, and Kun Gai. 2026.
\newblock \href {https://openreview.net/forum?id=NCN8oUsiNf} {Attention as a compass: Efficient exploration for process-supervised {RL} in reasoning models}.
\newblock In \emph{The Fourteenth International Conference on Learning Representations}.

\bibitem[{Naveed et~al.(2025)Naveed, Khan, Qiu, Saqib, Anwar, Usman, Akhtar, Barnes, and Mian}]{naveed2023comprehensive}
Humza Naveed, Asad~Ullah Khan, Shi Qiu, Muhammad Saqib, Saeed Anwar, Muhammad Usman, Naveed Akhtar, Nick Barnes, and Ajmal Mian. 2025.
\newblock \href {https://doi.org/10.1145/3744746} {A comprehensive overview of large language models}.
\newblock \emph{ACM Trans. Intell. Syst. Technol.}, 16(5).

\bibitem[{OpenAI et~al.(2024{\natexlab{a}})OpenAI, :, Hurst, Lerer, Goucher, Perelman, Ramesh, Clark, Ostrow, Welihinda, Hayes, Radford, Mądry, Baker-Whitcomb, Beutel, Borzunov, Carney, Chow, Kirillov, Nichol, Paino, Renzin, Passos, Kirillov, Christakis, Conneau, Kamali, Jabri, Moyer, Tam, Crookes, Tootoochian, Tootoonchian, Kumar, Vallone, Karpathy, Braunstein, Cann, Codispoti, Galu, Kondrich, Tulloch, Mishchenko, Baek, Jiang, Pelisse, Woodford, Gosalia, Dhar, Pantuliano, Nayak, Oliver, Zoph, Ghorbani, Leimberger, Rossen, Sokolowsky, Wang, Zweig, Hoover, Samic, McGrew, Spero, Giertler, Cheng, Lightcap, Walkin, Quinn, Guarraci, Hsu, Kellogg, Eastman, Lugaresi, Wainwright, Bassin, Hudson, Chu, Nelson, Li, Shern, Conger, Barette, Voss, Ding, Lu, Zhang, Beaumont, Hallacy, Koch, Gibson, Kim, Choi, McLeavey, Hesse, Fischer, Winter, Czarnecki, Jarvis, Wei, Koumouzelis, Sherburn, Kappler, Levin, Levy, Carr, Farhi, Mely, Robinson, Sasaki, Jin, Valladares, Tsipras, Li, Nguyen, Findlay, Oiwoh, Wong, Asdar, Proehl,
  Yang, Antonow, Kramer, Peterson, Sigler, Wallace, Brevdo, Mays, Khorasani, Such, Raso, Zhang, von Lohmann, Sulit, Goh, Oden, Salmon, Starace, Brockman, Salman, Bao, Hu, Wong, Wang, Schmidt, Whitney, Jun, Kirchner, de~Oliveira~Pinto, Ren, Chang, Chung, Kivlichan, O'Connell, O'Connell, Osband, Silber, Sohl, Okuyucu, Lan, Kostrikov, Sutskever, Kanitscheider, Gulrajani, Coxon, Menick, Pachocki, Aung, Betker, Crooks, Lennon, Kiros, Leike, Park, Kwon, Phang, Teplitz, Wei, Wolfe, Chen, Harris, Varavva, Lee, Shieh, Lin, Yu, Weng, Tang, Yu, Jang, Candela, Beutler, Landers, Parish, Heidecke, Schulman, Lachman, McKay, Uesato, Ward, Kim, Huizinga, Sitkin, Kraaijeveld, Gross, Kaplan, Snyder, Achiam, Jiao, Lee, Zhuang, Harriman, Fricke, Hayashi, Singhal, Shi, Karthik, Wood, Rimbach, Hsu, Nguyen, Gu-Lemberg, Button, Liu, Howe, Muthukumar, Luther, Ahmad, Kai, Itow, Workman, Pathak, Chen, Jing, Guy, Fedus, Zhou, Mamitsuka, Weng, McCallum, Held, Ouyang, Feuvrier, Zhang, Kondraciuk, Kaiser, Hewitt, Metz, Doshi, Aflak, Simens,
  Boyd, Thompson, Dukhan, Chen, Gray, Hudnall, Zhang, Aljubeh, Litwin, Zeng, Johnson, Shetty, Gupta, Shah, Yatbaz, Yang, Zhong, Glaese, Chen, Janner, Lampe, Petrov, Wu, Wang, Fradin, Pokrass, Castro, de~Castro, Pavlov, Brundage, Wang, Khan, Murati, Bavarian, Lin, Yesildal, Soto, Gimelshein, Cone, Staudacher, Summers, LaFontaine, Chowdhury, Ryder, Stathas, Turley, Tezak, Felix, Kudige, Keskar, Deutsch, Bundick, Puckett, Nachum, Okelola, Boiko, Murk, Jaffe, Watkins, Godement, Campbell-Moore, Chao, McMillan, Belov, Su, Bak, Bakkum, Deng, Dolan, Hoeschele, Welinder, Tillet, Pronin, Tillet, Dhariwal, Yuan, Dias, Lim, Arora, Troll, Lin, Lopes, Puri, Miyara, Leike, Gaubert, Zamani, Wang, Donnelly, Honsby, Smith, Sahai, Ramchandani, Huet, Carmichael, Zellers, Chen, Chen, Nigmatullin, Cheu, Jain, Altman, Schoenholz, Toizer, Miserendino, Agarwal, Culver, Ethersmith, Gray, Grove, Metzger, Hermani, Jain, Zhao, Wu, Jomoto, Wu, Shuaiqi, Xia, Phene, Papay, Narayanan, Coffey, Lee, Hall, Balaji, Broda, Stramer, Xu, Gogineni,
  Christianson, Sanders, Patwardhan, Cunninghman, Degry, Dimson, Raoux, Shadwell, Zheng, Underwood, Markov, Sherbakov, Rubin, Stasi, Kaftan, Heywood, Peterson, Walters, Eloundou, Qi, Moeller, Monaco, Kuo, Fomenko, Chang, Zheng, Zhou, Manassra, Sheu, Zaremba, Patil, Qian, Kim, Cheng, Zhang, He, Zhang, Jin, Dai, and Malkov}]{openai2024gpt4ocard}
OpenAI, :, Aaron Hurst, Adam Lerer, Adam~P. Goucher, Adam Perelman, Aditya Ramesh, Aidan Clark, AJ~Ostrow, Akila Welihinda, Alan Hayes, Alec Radford, Aleksander Mądry, Alex Baker-Whitcomb, Alex Beutel, Alex Borzunov, Alex Carney, Alex Chow, Alex Kirillov, and 401 others. 2024{\natexlab{a}}.
\newblock \href {https://arxiv.org/abs/2410.21276} {Gpt-4o system card}.
\newblock \emph{Preprint}, arXiv:2410.21276.

\bibitem[{OpenAI et~al.(2024{\natexlab{b}})OpenAI, :, Jaech, Kalai, Lerer, Richardson, El-Kishky, Low, Helyar, Madry, Beutel, Carney, Iftimie, Karpenko, Passos, Neitz, Prokofiev, Wei, Tam, Bennett, Kumar, Saraiva, Vallone, Duberstein, Kondrich, Mishchenko, Applebaum, Jiang, Nair, Zoph, Ghorbani, Rossen, Sokolowsky, Barak, McGrew, Minaiev, Hao, Baker, Houghton, McKinzie, Eastman, Lugaresi, Bassin, Hudson, Li, de~Bourcy, Voss, Shen, Zhang, Koch, Orsinger, Hesse, Fischer, Chan, Roberts, Kappler, Levy, Selsam, Dohan, Farhi, Mely, Robinson, Tsipras, Li, Oprica, Freeman, Zhang, Wong, Proehl, Cheung, Mitchell, Wallace, Ritter, Mays, Wang, Such, Raso, Leoni, Tsimpourlas, Song, von Lohmann, Sulit, Salmon, Parascandolo, Chabot, Zhao, Brockman, Leclerc, Salman, Bao, Sheng, Andrin, Bagherinezhad, Ren, Lightman, Chung, Kivlichan, O'Connell, Osband, Gilaberte, Akkaya, Kostrikov, Sutskever, Kofman, Pachocki, Lennon, Wei, Harb, Twore, Feng, Yu, Weng, Tang, Yu, Candela, Palermo, Parish, Heidecke, Hallman, Rizzo, Gordon,
  Uesato, Ward, Huizinga, Wang, Chen, Xiao, Singhal, Nguyen, Cobbe, Shi, Wood, Rimbach, Gu-Lemberg, Liu, Lu, Stone, Yu, Ahmad, Yang, Liu, Maksin, Ho, Fedus, Weng, Li, McCallum, Held, Kuhn, Kondraciuk, Kaiser, Metz, Boyd, Trebacz, Joglekar, Chen, Tintor, Meyer, Jones, Kaufer, Schwarzer, Shah, Yatbaz, Guan, Xu, Yan, Glaese, Chen, Lampe, Malek, Wang, Fradin, McClay, Pavlov, Wang, Wang, Murati, Bavarian, Rohaninejad, McAleese, Chowdhury, Chowdhury, Ryder, Tezak, Brown, Nachum, Boiko, Murk, Watkins, Chao, Ashbourne, Izmailov, Zhokhov, Dias, Arora, Lin, Lopes, Gaon, Miyara, Leike, Hwang, Garg, Brown, James, Shu, Cheu, Greene, Jain, Altman, Toizer, Toyer, Miserendino, Agarwal, Hernandez, Baker, McKinney, Yan, Zhao, Hu, Santurkar, Chaudhuri, Zhang, Fu, Papay, Lin, Balaji, Sanjeev, Sidor, Broda, Clark, Wang, Gordon, Sanders, Patwardhan, Sottiaux, Degry, Dimson, Zheng, Garipov, Stasi, Bansal, Creech, Peterson, Eloundou, Qi, Kosaraju, Monaco, Pong, Fomenko, Zheng, Zhou, McCabe, Zaremba, Dubois, Lu, Chen, Cha, Bai, He,
  Zhang, Wang, Shao, and Li}]{openai2024openaio1card}
OpenAI, :, Aaron Jaech, Adam Kalai, Adam Lerer, Adam Richardson, Ahmed El-Kishky, Aiden Low, Alec Helyar, Aleksander Madry, Alex Beutel, Alex Carney, Alex Iftimie, Alex Karpenko, Alex~Tachard Passos, Alexander Neitz, Alexander Prokofiev, Alexander Wei, Allison Tam, and 244 others. 2024{\natexlab{b}}.
\newblock \href {https://arxiv.org/abs/2412.16720} {Openai o1 system card}.
\newblock \emph{Preprint}, arXiv:2412.16720.

\bibitem[{Papineni et~al.(2002)Papineni, Roukos, Ward, and Zhu}]{papineni-etal-2002-bleu}
Kishore Papineni, Salim Roukos, Todd Ward, and Wei-Jing Zhu. 2002.
\newblock \href {https://doi.org/10.3115/1073083.1073135} {{B}leu: a method for automatic evaluation of machine translation}.
\newblock In \emph{Proceedings of the 40th Annual Meeting of the Association for Computational Linguistics}, pages 311--318, Philadelphia, Pennsylvania, USA. Association for Computational Linguistics.

\bibitem[{Qiu et~al.(2024)Qiu, He, Zhang, Li, and Lan}]{qiu-etal-2024-smile}
Huachuan Qiu, Hongliang He, Shuai Zhang, Anqi Li, and Zhenzhong Lan. 2024.
\newblock \href {https://doi.org/10.18653/v1/2024.findings-emnlp.34} {{SMILE}: Single-turn to multi-turn inclusive language expansion via {C}hat{GPT} for mental health support}.
\newblock In \emph{Findings of the Association for Computational Linguistics: EMNLP 2024}, pages 615--636, Miami, Florida, USA. Association for Computational Linguistics.

\bibitem[{Shao et~al.(2024)Shao, Wang, Zhu, Xu, Song, Bi, Zhang, Zhang, Li, Wu, and Guo}]{shao2024deepseekmath}
Zhihong Shao, Peiyi Wang, Qihao Zhu, Runxin Xu, Junxiao Song, Xiao Bi, Haowei Zhang, Mingchuan Zhang, Y.~K. Li, Y.~Wu, and Daya Guo. 2024.
\newblock \href {https://arxiv.org/abs/2402.03300} {Deepseekmath: Pushing the limits of mathematical reasoning in open language models}.
\newblock \emph{Preprint}, arXiv:2402.03300.

\bibitem[{Shen et~al.(2024)Shen, Li, Yang, Ni, Tao, Yu, Zheng, Xu, and Hu}]{10821773}
Hao Shen, Zihan Li, Minqiang Yang, Minghui Ni, Yongfeng Tao, Zhengyang Yu, Weihao Zheng, Chen Xu, and Bin Hu. 2024.
\newblock \href {https://doi.org/10.1109/BIBM62325.2024.10821773} {Are large language models possible to conduct cognitive behavioral therapy?}
\newblock In \emph{2024 IEEE International Conference on Bioinformatics and Biomedicine (BIBM)}, pages 3695--3700.

\bibitem[{Shen et~al.(2026)Shen, Fong, Jiang, Wang, Tang, Xu, Zhao, Xu, Liu, Hu, Dwyer, and Ge}]{shen2026psychethicsbenchevaluatinglargelanguage}
Yaling Shen, Stephanie Fong, Yiwen Jiang, Zimu Wang, Feilong Tang, Qingyang Xu, Xiangyu Zhao, Zhongxing Xu, Jiahe Liu, Jinpeng Hu, Dominic Dwyer, and Zongyuan Ge. 2026.
\newblock \href {https://doi.org/10.18653/v1/2026.findings-acl.1971} {{P}sych{E}thics{B}ench: Evaluating large language models against {A}ustralian mental health ethics}.
\newblock In \emph{Findings of the {A}ssociation for {C}omputational {L}inguistics: {ACL} 2026}, pages 39571--39589, San Diego, California, United States. Association for Computational Linguistics.

\bibitem[{Sheng et~al.(2025)Sheng, Zhang, Ye, Wu, Zhang, Zhang, Peng, Lin, and Wu}]{10.1145/3689031.3696075}
Guangming Sheng, Chi Zhang, Zilingfeng Ye, Xibin Wu, Wang Zhang, Ru~Zhang, Yanghua Peng, Haibin Lin, and Chuan Wu. 2025.
\newblock \href {https://doi.org/10.1145/3689031.3696075} {Hybridflow: A flexible and efficient rlhf framework}.
\newblock In \emph{Proceedings of the Twentieth European Conference on Computer Systems}, EuroSys '25, page 1279–1297, New York, NY, USA. Association for Computing Machinery.

\bibitem[{Shi et~al.(2026{\natexlab{a}})Shi, Sun, Liu, Yang, Fang, Sun, and Gu}]{shi2026reasoningtreesimprovingretrievalaugmented}
Yuling Shi, Maolin Sun, Zijun Liu, Mo~Yang, Yixiong Fang, Tianran Sun, and Xiaodong Gu. 2026{\natexlab{a}}.
\newblock \href {https://arxiv.org/abs/2601.11255} {Reasoning in trees: Improving retrieval-augmented generation for multi-hop question answering}.
\newblock \emph{Preprint}, arXiv:2601.11255.

\bibitem[{Shi et~al.(2026{\natexlab{b}})Shi, Xie, Sun, Chen, Zhang, Yun, Wan, Zhang, Lo, and Gu}]{shi2026codeocreffectivenessvisionlanguage}
Yuling Shi, Chaoxiang Xie, Zhensu Sun, Yeheng Chen, Chenxu Zhang, Longfei Yun, Chengcheng Wan, Hongyu Zhang, David Lo, and Xiaodong Gu. 2026{\natexlab{b}}.
\newblock \href {https://arxiv.org/abs/2602.01785} {Codeocr: On the effectiveness of vision language models in code understanding}.
\newblock \emph{Preprint}, arXiv:2602.01785.

\bibitem[{Song et~al.(2024)Song, Guo, Yang, Tang, and Wang}]{song2024emotional}
Peipei Song, Dan Guo, Xun Yang, Shengeng Tang, and Meng Wang. 2024.
\newblock \href {https://doi.org/10.1109/TIP.2024.3359045} {Emotional video captioning with vision-based emotion interpretation network}.
\newblock \emph{IEEE Transactions on Image Processing}, 33:1122--1135.

\bibitem[{Sorin et~al.(2024)Sorin, Brin, Barash, Konen, Charney, Nadkarni, and Klang}]{sorin2024large}
Vera Sorin, Dana Brin, Yiftach Barash, Eli Konen, Alexander Charney, Girish Nadkarni, and Eyal Klang. 2024.
\newblock \href {https://doi.org/10.2196/52597} {Large language models and empathy: Systematic review}.
\newblock \emph{J Med Internet Res}, 26:e52597.

\bibitem[{Tanana et~al.(2021)Tanana, Soma, Kuo, Bertagnolli, Dembe, Pace, Srikumar, Atkins, and Imel}]{tanana2021you}
Michael~J Tanana, Christina~S Soma, Patty~B Kuo, Nicolas~M Bertagnolli, Aaron Dembe, Brian~T Pace, Vivek Srikumar, David~C Atkins, and Zac~E Imel. 2021.
\newblock \href {https://doi.org/10.3758/s13428-020-01531-z} {How do you feel? using natural language processing to automatically rate emotion in psychotherapy}.
\newblock \emph{Behavior research methods}, 53(5):2069--2082.

\bibitem[{Team(2024{\natexlab{a}})}]{2024EmoLLM}
EmoLLM Team. 2024{\natexlab{a}}.
\newblock Emollm: Reinventing mental health support with large language models.
\newblock \url{https://github.com/SmartFlowAI/EmoLLM}.

\bibitem[{Team et~al.(2024)Team, Georgiev, Lei, Burnell, Bai, Gulati, Tanzer, Vincent, Pan, Wang, Mariooryad, Ding, Geng, Alcober, Frostig, Omernick, Walker, Paduraru, Sorokin, Tacchetti, Gaffney, Daruki, Sercinoglu, Gleicher, Love, Voigtlaender, Jain, Surita, Mohamed, Blevins, Ahn, Zhu, Kawintiranon, Firat, Gu, Zhang, Rahtz, Faruqui, Clay, Gilmer, Co-Reyes, Penchev, Zhu, Morioka, Hui, Haridasan, Campos, Mahdieh, Guo, Hassan, Kilgour, Vezer, Cheng, de~Liedekerke, Goyal, Barham, Strouse, Noury, Adler, Sundararajan, Vikram, Lepikhin, Paganini, Garcia, Yang, Valter, Trebacz, Vodrahalli, Asawaroengchai, Ring, Kalb, Soares, Brahma, Steiner, Yu, Mentzer, He, Gonzalez, Xu, Kaufman, Shafey, Oh, Hennigan, van~den Driessche, Odoom, Lucic, Roelofs, Lall, Marathe, Chan, Ontanon, He, Teplyashin, Lai, Crone, Damoc, Ho, Riedel, Lenc, Yeh, Chowdhery, Xu, Kazemi, Amid, Petrushkina, Swersky, Khodaei, Chen, Larkin, Pinto, Yan, Badia, Patil, Hansen, Orr, Arnold, Grimstad, Dai, Douglas, Sinha, Yadav, Chen, Gribovskaya, Austin,
  Zhao, Patel, Komarek, Austin, Borgeaud, Friso, Goyal, Caine, Cao, Chung, Lamm, Barth-Maron, Kagohara, Olszewska, Chen, Shivakumar, Agarwal, Godhia, Rajwar, Snaider, Dotiwalla, Liu, Barua, Ungureanu, Zhang, Batsaikhan, Wirth, Qin, Danihelka, Doshi, Chadwick, Chen, Jain, Le, Kar, Gurumurthy, Li, Sang, Liu, Lamprou, Munoz, Lintz, Mehta, Howard, Reynolds, Aroyo, Wang, Blanco, Cassirer, Griffith, Das, Lee, Sygnowski, Fisher, Besley, Powell, Ahmed, Paulus, Reitter, Borsos, Joshi, Pope, Hand, Selo, Jain, Sethi, Goel, Makino, May, Yang, Schalkwyk, Butterfield, Hauth, Goldin, Hawkins, Senter, Brin, Woodman, Ritter, Noland, Giang, Bolina, Lee, Blyth, Mackinnon, Reid, Sarvana, Silver, Chen, Wang, Maggiore, Chang, Attaluri, Thornton, Chiu, Bunyan, Levine, Chung, Eltyshev, Si, Lillicrap, Brady, Aggarwal, Wu, Xu, McIlroy, Badola, Sandhu, Moreira, Stokowiec, Hemsley, Li, Tudor, Shyam, Rahimtoroghi, Haykal, Sprechmann, Zhou, Mincu, Li, Addanki, Krishna, Wu, Frechette, Eyal, Dafoe, Lacey, Whang, Avrahami, Zhang, Taropa,
  Lin, Toyama, Rutherford, Sano, Choe, Tomala, Safranek-Shrader, Kassner, Pajarskas, Harvey, Sechrist, Fortunato, Lyu, Elsayed, Kuang, Lottes, Chu, Jia, Chen, Humphreys, Baumli, Tao, Samuel, dos Santos, Andreassen, Rakićević, Grewe, Kumar, Winkler, Caton, Brock, Dalmia, Sheahan, Barr, Miao, Natsev, Devlin, Behbahani, Prost, Sun, Myaskovsky, Pillai, Hurt, Lazaridou, Xiong, Zheng, Pardo, Li, Horgan, Stanton, Ambar, Xia, Lince, Wang, Mustafa, Webson, Lee, Anil, Wicke, Dozat, Sinha, Piqueras, Dabir, Upadhyay, Boral, Hendricks, Fry, Djolonga, Su, Walker, Labanowski, Huang, Misra, Chen, Skerry-Ryan, Singh, Rijhwani, Yu, Castro-Ros, Changpinyo, Datta, Bagri, Hrafnkelsson, Maggioni, Zheng, Sulsky, Hou, Paine, Yang, Riesa, Rogozinska, Marcus, Badawy, Zhang, Wang, Miller, Greer, Sjos, Nova, Zen, Chaabouni, Rosca, Jiang, Chen, Liu, Sainath, Krikun, Polozov, Lespiau, Newlan, Cankara, Kwak, Xu, Chen, Coenen, Meyer, Tsihlas, Ma, Gottweis, Xing, Gu, Miao, Frank, Cankara, Ganapathy, Dasgupta, Hughes-Fitt, Chen, Reid, Rong,
  Fan, van Amersfoort, Zhuang, Cohen, Gu, Mohananey, Ilic, Tobin, Wieting, Bortsova, Thacker, Wang, Caveness, Chiu, Sezener, Kaskasoli, Baker, Millican, Elhawaty, Aisopos, Lebsack, Byrd, Dai, Jia, Wiethoff, Davoodi, Weston, Yagati, Ahuja, Gao, Pundak, Zhang, Azzam, Sim, Caelles, Keeling, Sharma, Swing, Li, Liu, Bostock, Bansal, Nado, Anand, Lipschultz, Karmarkar, Proleev, Ittycheriah, Yeganeh, Polovets, Faust, Sun, Rrustemi, Li, Shivanna, Liu, Welty, Lebron, Baddepudi, Krause, Parisotto, Soricut, Xu, Bloxwich, Johnson, Neyshabur, Mao-Jones, Wang, Ramasesh, Abbas, Guez, Segal, Nguyen, Svensson, Hou, York, Milan, Bridgers, Gworek, Tagliasacchi, Lee-Thorp, Chang, Guseynov, Hartman, Kwong, Zhao, Kashem, Cole, Miech, Tanburn, Phuong, Pavetic, Cevey, Comanescu, Ives, Yang, Du, Li, Zhang, Iinuma, Hu, Roy, Bijwadia, Zhu, Martins, Saputro, Gergely, Zheng, Jia, Antonoglou, Sadovsky, Gu, Bi, Andreev, Samangooei, Khan, Kocisky, Filos, Kumar, Bishop, Yu, Hodkinson, Mittal, Shah, Moufarek, Cheng, Bloniarz, Lee, Pejman,
  Michel, Spencer, Feinberg, Xiong, Savinov, Smith, Shakeri, Tran, Chesus, Bohnet, Tucker, von Glehn, Muir, Mao, Kazawa, Slone, Soparkar, Shrivastava, Cobon-Kerr, Sharman, Pavagadhi, Araya, Misiunas, Ghelani, Laskin, Barker, Li, Briukhov, Houlsby, Glaese, Lakshminarayanan, Schucher, Tang, Collins, Lim, Feng, Recasens, Lai, Magni, Cao, Siddhant, Ashwood, Orbay, Dehghani, Brennan, He, Xu, Gao, Saroufim, Molloy, Wu, Arnold, Chang, Schrittwieser, Buchatskaya, Radpour, Polacek, Giordano, Bapna, Tokumine, Hellendoorn, Sottiaux, Cogan, Severyn, Saleh, Thakoor, Shefey, Qiao, Gaba, yiin Chang, Swanson, Zhang, Lee, Rubenstein, Song, Kwiatkowski, Koop, Kannan, Kao, Schuh, Stjerngren, Ghiasi, Gibson, Vilnis, Yuan, Ferreira, Kamath, Klimenko, Franko, Xiao, Bhattacharya, Patel, Wang, Morris, Strudel, Sharma, Choy, Hashemi, Landon, Finkelstein, Jhakra, Frye, Barnes, Mauger, Daun, Baatarsukh, Tung, Farhan, Michalewski, Viola, de~Chaumont~Quitry, Lan, Hudson, Wang, Fischer, Zheng, White, Dragan, baptiste Alayrac, Ni, Pritzel,
  Iwanicki, Isard, Bulanova, Zilka, Dyer, Sachan, Srinivasan, Muckenhirn, Cai, Mandhane, Tariq, Rae, Wang, Ayoub, FitzGerald, Zhao, Han, Alberti, Garrette, Krishnakumar, Gimenez, Levskaya, Sohn, Matak, Iturrate, Chang, Xiang, Cao, Ranka, Brown, Hutter, Mirrokni, Chen, Yao, Egyed, Galilee, Liechty, Kallakuri, Palmer, Ghemawat, Liu, Tao, Thornton, Green, Jasarevic, Lin, Cotruta, Tan, Fiedel, Yu, Chi, Neitz, Heitkaemper, Sinha, Zhou, Sun, Kaed, Hulse, Mishra, Georgaki, Kudugunta, Farabet, Shafran, Vlasic, Tsitsulin, Ananthanarayanan, Carin, Su, Sun, V, Carvajal, Broder, Comsa, Repina, Wong, Chen, Hawkins, Filonov, Loher, Hirnschall, Wang, Ye, Burns, Cate, Wright, Piccinini, Zhang, Lin, Gog, Kulizhskaya, Sreevatsa, Song, Cobo, Iyer, Tekur, Garrido, Xiao, Kemp, Zheng, Li, Agarwal, Ngani, Goshvadi, Santamaria-Fernandez, Fica, Chen, Gorgolewski, Sun, Garg, Ye, Eslami, Hua, Simon, Joshi, Kim, Tenney, Potluri, Thiet, Yuan, Luisier, Chronopoulou, Scellato, Srinivasan, Chen, Koverkathu, Dalibard, Xu, Saeta, Anderson,
  Sellam, Fernando, Huot, Jung, Varadarajan, Quinn, Raul, Le, Habalov, Clark, Jalan, Bullard, Singhal, Luong, Wang, Rajayogam, Eisenschlos, Jia, Finchelstein, Yakubovich, Balle, Fink, Agarwal, Li, Dvijotham, Pal, Kang, Konzelmann, Beattie, Dousse, Wu, Crocker, Elkind, Jonnalagadda, Lee, Holtmann-Rice, Kallarackal, Liu, Vnukov, Vats, Invernizzi, Jafari, Zhou, Taylor, Prendki, Wu, Eccles, Liu, Kopparapu, Beaufays, Angermueller, Marzoca, Sarcar, Dib, Stanway, Perbet, Trdin, Sterneck, Khorlin, Li, Wu, Goenka, Madras, Goldshtein, Gierke, Zhou, Liu, Liang, White, Li, Singh, Bahargam, Epstein, Basu, Lao, Ozturel, Crous, Zhai, Lu, Tung, Gaur, Walton, Dixon, Zhang, Globerson, Uy, Bolt, Wiles, Nasr, Shumailov, Selvi, Piccinno, Aguilar, McCarthy, Khalman, Shukla, Galic, Carpenter, Villela, Zhang, Richardson, Martens, Bosnjak, Belle, Seibert, Alnahlawi, McWilliams, Singh, Louis, Ding, Popovici, Simicich, Knight, Mehta, Gupta, Shi, Fatehi, Mitrovic, Grills, Pagadora, Munkhdalai, Petrova, Eisenbud, Zhang, Yates, Mittal,
  Tripuraneni, Assael, Brovelli, Jain, Velimirovic, Akbulut, Mu, Macherey, Kumar, Xu, Qureshi, Comanici, Wiesner, Gong, Ruddock, Bauer, Felt, GP, Arnab, Zelle, Rothfuss, Rosgen, Shenoy, Seybold, Li, Mudigonda, Erdogan, Xia, Simsa, Michi, Yao, Yew, Kan, Caswell, Radebaugh, Elisseeff, Valenzuela, McKinney, Paterson, Cui, Latorre-Chimoto, Kim, Zeng, Durden, Ponnapalli, Sosea, Choquette-Choo, Manyika, Robenek, Vashisht, Pereira, Lam, Velic, Owusu-Afriyie, Lee, Bolukbasi, Parrish, Lu, Park, Venkatraman, Talbert, Rosique, Cheng, Sozanschi, Paszke, Kumar, Austin, Li, Salama, Perz, Kim, Dukkipati, Baryshnikov, Kaplanis, Sheng, Chervonyi, Unlu, de~Las~Casas, Askham, Tunyasuvunakool, Gimeno, Poder, Kwak, Miecnikowski, Mirrokni, Dimitriev, Parisi, Liu, Tsai, Shevlane, Kouridi, Garmon, Goedeckemeyer, Brown, Vijayakumar, Elqursh, Jazayeri, Huang, Carthy, Hoover, Kim, Kumar, Chen, Biles, Bingham, Rosen, Wang, Tan, Engel, Pongetti, de~Cesare, Hwang, Yu, Pullman, Narayanan, Levin, Gopal, Li, Aharoni, Trinh, Lo, Casagrande,
  Vij, Matthey, Ramadhana, Matthews, Carey, Johnson, Goranova, Shah, Ashraf, Dasgupta, Larsen, Wang, Vuyyuru, Jiang, Ijazi, Osawa, Smith, Boppana, Bilal, Koizumi, Xu, Altun, Shabat, Bariach, Korchemniy, Choo, Ronneberger, Iwuanyanwu, Zhao, Soergel, Hsieh, Cai, Iqbal, Sundermeyer, Chen, Bursztein, Malaviya, Biadsy, Shroff, Dhillon, Latkar, Dyer, Forbes, Nicosia, Nikolaev, Greene, Georgiev, Wang, Martin, Sedghi, Zhang, Banzal, Fritz, Rao, Wang, Zhang, Patraucean, Du, Mordatch, Jurin, Liu, Dubey, Mohan, Nowakowski, Ion, Wei, Tojo, Raad, Hudson, Keshava, Agrawal, Ramirez, Wu, Nguyen, Liu, Sewak, Petrini, Choi, Philips, Wang, Bica, Garg, Wilkiewicz, Agrawal, Li, Guo, Xue, Shaik, Leach, Khan, Wiesinger, Jerome, Chakladar, Wang, Ornduff, Abu, Ghaffarkhah, Wainwright, Cortes, Liu, Maynez, Terzis, Samangouei, Mansour, Kępa, Aubet, Algymr, Banica, Weisz, Orban, Senges, Andrejczuk, Geller, Santo, Anklin, Merey, Baeuml, Strohman, Bai, Petrov, Wu, Hassabis, Kavukcuoglu, Dean, and Vinyals}]{geminiteam2024}
Gemini Team, Petko Georgiev, Ving~Ian Lei, Ryan Burnell, Libin Bai, Anmol Gulati, Garrett Tanzer, Damien Vincent, Zhufeng Pan, Shibo Wang, Soroosh Mariooryad, Yifan Ding, Xinyang Geng, Fred Alcober, Roy Frostig, Mark Omernick, Lexi Walker, Cosmin Paduraru, Christina Sorokin, and 1118 others. 2024.
\newblock \href {https://arxiv.org/abs/2403.05530} {Gemini 1.5: Unlocking multimodal understanding across millions of tokens of context}.
\newblock \emph{Preprint}, arXiv:2403.05530.

\bibitem[{Team et~al.(2025)Team, Xiao, Li, Han, Bai, Cai, Chen, Chen, Cong, Cui, Ding, Fan, Fang, Fu, Guan, Guan, Guo, Han, He, Huang, Ji, Kong, Li, Li, Li, Li, Li, Li, Li, Liu, Lin, Lin, Long, Lu, Lu, Luo, Lyu, Ou, Pan, Pu, Qu, Shi, Song, Su, Su, Sun, Sun, Tang, Wang, Wang, Wang, Wang, Wang, Wu, Xiao, Xie, Xie, Xu, Yan, Yuan, Zhang, Zhang, Zhang, Zhang, Zhang, Zhang, Zhao, Zhao, Zhao, Zhao, Zheng, Zhou, Zhou, Zhou, Zhou, Zhou, Zhou, Zhou, Liu, Zeng, Jia, Li, and Sun}]{team2025minicpm4}
MiniCPM Team, Chaojun Xiao, Yuxuan Li, Xu~Han, Yuzhuo Bai, Jie Cai, Haotian Chen, Wentong Chen, Xin Cong, Ganqu Cui, Ning Ding, Shengda Fan, Yewei Fang, Zixuan Fu, Wenyu Guan, Yitong Guan, Junshao Guo, Yufeng Han, Bingxiang He, and 64 others. 2025.
\newblock \href {https://arxiv.org/abs/2506.07900} {Minicpm4: Ultra-efficient llms on end devices}.
\newblock \emph{Preprint}, arXiv:2506.07900.

\bibitem[{Team(2024{\natexlab{b}})}]{qwen2.5}
Qwen Team. 2024{\natexlab{b}}.
\newblock \href {https://qwenlm.github.io/blog/qwen2.5/} {Qwen2.5: A party of foundation models}.

\bibitem[{Wei et~al.(2022)Wei, Wang, Schuurmans, Bosma, ichter, Xia, Chi, Le, and Zhou}]{NEURIPS2022_9d560961}
Jason Wei, Xuezhi Wang, Dale Schuurmans, Maarten Bosma, brian ichter, Fei Xia, Ed~Chi, Quoc~V Le, and Denny Zhou. 2022.
\newblock \href {https://proceedings.neurips.cc/paper_files/paper/2022/file/9d5609613524ecf4f15af0f7b31abca4-Paper-Conference.pdf} {Chain-of-thought prompting elicits reasoning in large language models}.
\newblock In \emph{Advances in Neural Information Processing Systems}, volume~35, pages 24824--24837. Curran Associates, Inc.

\bibitem[{Wolohan et~al.(2018)Wolohan, Hiraga, Mukherjee, Sayyed, and Millard}]{wolohan2018detecting}
JT~Wolohan, Misato Hiraga, Atreyee Mukherjee, Zeeshan~Ali Sayyed, and Matthew Millard. 2018.
\newblock Detecting linguistic traces of depression in topic-restricted text: Attending to self-stigmatized depression with nlp.
\newblock In \emph{Proceedings of the first international workshop on language cognition and computational models}, pages 11--21.

\bibitem[{Wu et~al.(2025)Wu, Huang, and Lu}]{wu2025psychological}
Shurui Wu, Xinyi Huang, and Dingxin Lu. 2025.
\newblock \href {https://doi.org/10.1145/3733006.3733032} {Psychological health knowledge-enhanced llm-based social network crisis intervention text transfer recognition method}.
\newblock In \emph{Proceedings of the 2025 International Conference on Health Big Data}, HBD '25, page 156–161, New York, NY, USA. Association for Computing Machinery.

\bibitem[{Xiao et~al.(2024)Xiao, Xie, Kuang, Liu, Yang, Peng, Han, and Huang}]{xiao-etal-2024-healme}
Mengxi Xiao, Qianqian Xie, Ziyan Kuang, Zhicheng Liu, Kailai Yang, Min Peng, Weiguang Han, and Jimin Huang. 2024.
\newblock \href {https://doi.org/10.18653/v1/2024.acl-long.93} {{H}eal{M}e: Harnessing cognitive reframing in large language models for psychotherapy}.
\newblock In \emph{Proceedings of the 62nd Annual Meeting of the Association for Computational Linguistics (Volume 1: Long Papers)}, pages 1707--1725, Bangkok, Thailand. Association for Computational Linguistics.

\bibitem[{Xiao et~al.(2025)Xiao, Yang, Zhao, Zhang, Kuang, Liu, Han, Liao, Huang, Hu, Peng, Xie, and Ananiadou}]{xiao2025mentrasuiteposttraininglargelanguage}
Mengxi Xiao, Kailai Yang, Pengde Zhao, Enze Zhang, Ziyan Kuang, Zhiwei Liu, Weiguang Han, Shu Liao, Lianting Huang, Jinpeng Hu, Min Peng, Qianqian Xie, and Sophia Ananiadou. 2025.
\newblock \href {https://arxiv.org/abs/2512.09636} {Mentrasuite: Post-training large language models for mental health reasoning and assessment}.
\newblock \emph{Preprint}, arXiv:2512.09636.

\bibitem[{Xie et~al.(2025)Xie, Chen, Xing, Lin, and Xu}]{xie-etal-2025-psydt}
Haojie Xie, Yirong Chen, Xiaofen Xing, Jingkai Lin, and Xiangmin Xu. 2025.
\newblock \href {https://aclanthology.org/2025.acl-long.55/} {{P}sy{DT}: Using {LLM}s to construct the digital twin of psychological counselor with personalized counseling style for psychological counseling}.
\newblock In \emph{Proceedings of the 63rd Annual Meeting of the Association for Computational Linguistics (Volume 1: Long Papers)}, pages 1081--1115, Vienna, Austria. Association for Computational Linguistics.

\bibitem[{Xu et~al.(2026)Xu, Hu, Song, Duan, and Yang}]{10.1145/3774904.3792594}
Yangyang Xu, Jinpeng Hu, Peipei Song, Zhangling Duan, and Xun Yang. 2026.
\newblock \href {https://doi.org/10.1145/3774904.3792594} {From social media to psychological scale: An adaptive framework with two-hop retrieval for depression screening}.
\newblock In \emph{Proceedings of the ACM Web Conference 2026}, WWW '26, page 4817–4828, New York, NY, USA. Association for Computing Machinery.

\bibitem[{Xu et~al.(2025)Xu, Hu, Zhao, Duan, Sun, and Yang}]{xu-etal-2025-multiagentesc}
Yangyang Xu, Jinpeng Hu, Zhuoer Zhao, Zhangling Duan, Xiao Sun, and Xun Yang. 2025.
\newblock \href {https://doi.org/10.18653/v1/2025.emnlp-main.232} {{M}ulti{A}gent{ESC}: A {LLM}-based multi-agent collaboration framework for emotional support conversation}.
\newblock In \emph{Proceedings of the 2025 Conference on Empirical Methods in Natural Language Processing}, pages 4665--4681, Suzhou, China. Association for Computational Linguistics.

\bibitem[{Yang et~al.(2025)Yang, Li, Yang, Zhang, Hui, Zheng, Yu, Gao, Huang, Lv, Zheng, Liu, Zhou, Huang, Hu, Ge, Wei, Lin, Tang, Yang, Tu, Zhang, Yang, Yang, Zhou, Zhou, Lin, Dang, Bao, Yang, Yu, Deng, Li, Xue, Li, Zhang, Wang, Zhu, Men, Gao, Liu, Luo, Li, Tang, Yin, Ren, Wang, Zhang, Ren, Fan, Su, Zhang, Zhang, Wan, Liu, Wang, Cui, Zhang, Zhou, and Qiu}]{yang2025qwen3technicalreport}
An~Yang, Anfeng Li, Baosong Yang, Beichen Zhang, Binyuan Hui, Bo~Zheng, Bowen Yu, Chang Gao, Chengen Huang, Chenxu Lv, Chujie Zheng, Dayiheng Liu, Fan Zhou, Fei Huang, Feng Hu, Hao Ge, Haoran Wei, Huan Lin, Jialong Tang, and 41 others. 2025.
\newblock \href {https://arxiv.org/abs/2505.09388} {Qwen3 technical report}.
\newblock \emph{Preprint}, arXiv:2505.09388.

\bibitem[{Yang et~al.(2024)Yang, Wang, Chen, Wang, Pu, Gao, Huang, Song, and Huang}]{yang-etal-2024-psychogat}
Qisen Yang, Zekun Wang, Honghui Chen, Shenzhi Wang, Yifan Pu, Xin Gao, Wenhao Huang, Shiji Song, and Gao Huang. 2024.
\newblock \href {https://doi.org/10.18653/v1/2024.acl-long.779} {{P}sycho{GAT}: A novel psychological measurement paradigm through interactive fiction games with {LLM} agents}.
\newblock In \emph{Proceedings of the 62nd Annual Meeting of the Association for Computational Linguistics (Volume 1: Long Papers)}, pages 14470--14505, Bangkok, Thailand. Association for Computational Linguistics.

\bibitem[{Yao et~al.(2023)Yao, Yu, Zhao, Shafran, Griffiths, Cao, and Narasimhan}]{NEURIPS2023_271db992}
Shunyu Yao, Dian Yu, Jeffrey Zhao, Izhak Shafran, Tom Griffiths, Yuan Cao, and Karthik Narasimhan. 2023.
\newblock \href {https://proceedings.neurips.cc/paper_files/paper/2023/file/271db9922b8d1f4dd7aaef84ed5ac703-Paper-Conference.pdf} {Tree of thoughts: Deliberate problem solving with large language models}.
\newblock In \emph{Advances in Neural Information Processing Systems}, volume~36, pages 11809--11822. Curran Associates, Inc.

\bibitem[{Ye et~al.(2025)Ye, Xiang, Zhang, and Zong}]{ye-etal-2025-sweetiechat}
Jing Ye, Lu~Xiang, Yaping Zhang, and Chengqing Zong. 2025.
\newblock \href {https://aclanthology.org/2025.coling-main.312/} {{S}weetie{C}hat: A strategy-enhanced role-playing framework for diverse scenarios handling emotional support agent}.
\newblock In \emph{Proceedings of the 31st International Conference on Computational Linguistics}, pages 4646--4669, Abu Dhabi, UAE. Association for Computational Linguistics.

\bibitem[{Yu et~al.(2025)Yu, Zhang, Zhu, Yuan, Zuo, YuYue, Dai, Fan, Liu, Liu, Liu, Liu, Lin, Lin, Ma, Sheng, Tong, Zhang, Zhang, Zhang, Zhang, Zhu, Zhu, Chen, Chen, Wang, Yu, Song, Wei, Zhou, Liu, Ma, Zhang, Yan, Wu, and Wang}]{yu2025dapo}
Qiying Yu, Zheng Zhang, Ruofei Zhu, Yufeng Yuan, Xiaochen Zuo, YuYue, Weinan Dai, Tiantian Fan, Gaohong Liu, Juncai Liu, LingJun Liu, Xin Liu, Haibin Lin, Zhiqi Lin, Bole Ma, Guangming Sheng, Yuxuan Tong, Chi Zhang, Mofan Zhang, and 17 others. 2025.
\newblock \href {https://openreview.net/forum?id=2a36EMSSTp} {{DAPO}: An open-source {LLM} reinforcement learning system at scale}.
\newblock In \emph{The Thirty-ninth Annual Conference on Neural Information Processing Systems}.

\bibitem[{Zelikman et~al.(2022)Zelikman, Wu, Mu, and Goodman}]{NEURIPS2022_639a9a17}
Eric Zelikman, Yuhuai Wu, Jesse Mu, and Noah Goodman. 2022.
\newblock \href {https://proceedings.neurips.cc/paper_files/paper/2022/file/639a9a172c044fbb64175b5fad42e9a5-Paper-Conference.pdf} {Star: Bootstrapping reasoning with reasoning}.
\newblock In \emph{Advances in Neural Information Processing Systems}, volume~35, pages 15476--15488. Curran Associates, Inc.

\bibitem[{Zhang et~al.(2024)Zhang, Li, Tan, Yang, Zhu, Yang, Zhao, Ye, Li, and Hu}]{zhang-etal-2024-cpsycoun}
Chenhao Zhang, Renhao Li, Minghuan Tan, Min Yang, Jingwei Zhu, Di~Yang, Jiahao Zhao, Guancheng Ye, Chengming Li, and Xiping Hu. 2024.
\newblock \href {https://doi.org/10.18653/v1/2024.findings-acl.830} {{CP}sy{C}oun: A report-based multi-turn dialogue reconstruction and evaluation framework for {C}hinese psychological counseling}.
\newblock In \emph{Findings of the Association for Computational Linguistics: ACL 2024}, pages 13947--13966, Bangkok, Thailand. Association for Computational Linguistics.

\bibitem[{Zhang et~al.(2025{\natexlab{a}})Zhang, Li, Bao, Gao, Xiao, Zhang, Zhang, Huang, Wu, Wang, and Xu}]{zhang2025hyperadaloraacceleratinglorarank}
Hao Zhang, Zhenjia Li, Runfeng Bao, Yifan Gao, Xi~Xiao, Heng Zhang, Shuyang Zhang, Bo~Huang, Yuhang Wu, Tianyang Wang, and Hao Xu. 2025{\natexlab{a}}.
\newblock \href {https://arxiv.org/abs/2510.02630} {Hyperadalora: Accelerating lora rank allocation during training via hypernetworks without sacrificing performance}.
\newblock \emph{Preprint}, arXiv:2510.02630.

\bibitem[{Zhang et~al.(2023)Zhang, Chen, Jiang, Yu, Chen, Chen, Li, Wu, Zhiyi, Xiao, Wan, Wang, and Li}]{zhang-etal-2023-huatuogpt}
Hongbo Zhang, Junying Chen, Feng Jiang, Fei Yu, Zhihong Chen, Guiming Chen, Jianquan Li, Xiangbo Wu, Zhang Zhiyi, Qingying Xiao, Xiang Wan, Benyou Wang, and Haizhou Li. 2023.
\newblock \href {https://doi.org/10.18653/v1/2023.findings-emnlp.725} {{H}uatuo{GPT}, towards taming language model to be a doctor}.
\newblock In \emph{Findings of the Association for Computational Linguistics: EMNLP 2023}, pages 10859--10885, Singapore. Association for Computational Linguistics.

\bibitem[{Zhang et~al.(2025{\natexlab{b}})Zhang, He, Ma, Song, He, Zhang, Qiu, Zhou, Li, Dai, Xu, and Lan}]{zhang2025conceptpsy}
Junlei Zhang, Hongliang He, Lizhi Ma, Nirui Song, Shuyuan He, Shuai Zhang, Huachuan Qiu, Zhanchao Zhou, Anqi Li, Yong Dai, Renjun Xu, and Zhenzhong Lan. 2025{\natexlab{b}}.
\newblock \href {https://doi.org/10.1016/j.neucom.2025.130070} {Conceptpsy: A comprehensive benchmark suite for hierarchical psychological concept understanding in llms}.
\newblock \emph{Neurocomputing}, 637:130070.

\bibitem[{Zhang et~al.(2026{\natexlab{a}})Zhang, Cheng, Ye, and Huang}]{zhang2026cogevolutionhumanlikegenerativeeducational}
Wei Zhang, Yihang Cheng, Zhirong Ye, and Kezhen Huang. 2026{\natexlab{a}}.
\newblock \href {https://arxiv.org/abs/2604.14786} {Cogevolution: A human-like generative educational agent to simulate student's cognitive evolution}.
\newblock \emph{Preprint}, arXiv:2604.14786.

\bibitem[{Zhang et~al.(2026{\natexlab{b}})Zhang, Zhang, Chen, Yu, Yang, and Song}]{zhang2026logicalphasetransitionsunderstanding}
Xinglang Zhang, Yunyao Zhang, ZeLiang Chen, Junqing Yu, Wei Yang, and Zikai Song. 2026{\natexlab{b}}.
\newblock \href {https://arxiv.org/abs/2601.02902} {Logical phase transitions: Understanding collapse in llm logical reasoning}.
\newblock \emph{Preprint}, arXiv:2601.02902.

\bibitem[{Zhang et~al.(2026{\natexlab{c}})Zhang, Zhang, Sheng, Li, Yu, Chen, Yang, and Song}]{zhang2026semanticawarelogicalreasoningsemiotic}
Yunyao Zhang, Xinglang Zhang, Junxi Sheng, Wenbing Li, Junqing Yu, Yi-Ping~Phoebe Chen, Wei Yang, and Zikai Song. 2026{\natexlab{c}}.
\newblock \href {https://arxiv.org/abs/2509.24765} {Semantic-aware logical reasoning via a semiotic framework}.
\newblock \emph{Preprint}, arXiv:2509.24765.

\bibitem[{Zhao et~al.(2025)Zhao, Zhu, Tan, Yang, Li, Di, Zhang, Ye, Li, Hu, and Wong}]{zhao-etal-2025-cpsyexam}
Jiahao Zhao, Jingwei Zhu, Minghuan Tan, Min Yang, Renhao Li, Yang Di, Chenhao Zhang, Guancheng Ye, Chengming Li, Xiping Hu, and Derek~F. Wong. 2025.
\newblock \href {https://aclanthology.org/2025.coling-main.745/} {{CP}sy{E}xam: A {C}hinese benchmark for evaluating psychology using examinations}.
\newblock In \emph{Proceedings of the 31st International Conference on Computational Linguistics}, pages 11248--11260, Abu Dhabi, UAE. Association for Computational Linguistics.

\bibitem[{Zheng et~al.(2023)Zheng, Sabour, Wen, Zhang, and Huang}]{zheng-etal-2023-augesc}
Chujie Zheng, Sahand Sabour, Jiaxin Wen, Zheng Zhang, and Minlie Huang. 2023.
\newblock \href {https://doi.org/10.18653/v1/2023.findings-acl.99} {{A}ug{ESC}: Dialogue augmentation with large language models for emotional support conversation}.
\newblock In \emph{Findings of the Association for Computational Linguistics: ACL 2023}, pages 1552--1568, Toronto, Canada. Association for Computational Linguistics.

\bibitem[{Zheng et~al.(2024)Zheng, Zhang, Zhang, Ye, and Luo}]{zheng-etal-2024-llamafactory}
Yaowei Zheng, Richong Zhang, Junhao Zhang, Yanhan Ye, and Zheyan Luo. 2024.
\newblock \href {https://doi.org/10.18653/v1/2024.acl-demos.38} {{L}lama{F}actory: Unified efficient fine-tuning of 100+ language models}.
\newblock In \emph{Proceedings of the 62nd Annual Meeting of the Association for Computational Linguistics (Volume 3: System Demonstrations)}, pages 400--410, Bangkok, Thailand. Association for Computational Linguistics.

\bibitem[{Zhou et~al.(2026)Zhou, Li, Cheng, Fan, and Cheng}]{zhou2026lookinwardexploreoutward}
Yixiao Zhou, Yang Li, Dongzhou Cheng, Hehe Fan, and Yu~Cheng. 2026.
\newblock \href {https://arxiv.org/abs/2602.13035} {Look inward to explore outward: Learning temperature policy from llm internal states via hierarchical rl}.
\newblock \emph{Preprint}, arXiv:2602.13035.

\bibitem[{Zhu et~al.(2025)Zhu, Chen, Dou, Li, Guo, Chen, and Zhang}]{zhu2025dianjin}
Jie Zhu, Qian Chen, Huaixia Dou, Junhui Li, Lifan Guo, Feng Chen, and Chi Zhang. 2025.
\newblock \href {https://arxiv.org/abs/2504.15716} {Dianjin-r1: Evaluating and enhancing financial reasoning in large language models}.
\newblock \emph{Preprint}, arXiv:2504.15716.

\end{thebibliography}

\appendix

\begin{figure*}[t]
\centering
\includegraphics[width=\textwidth]{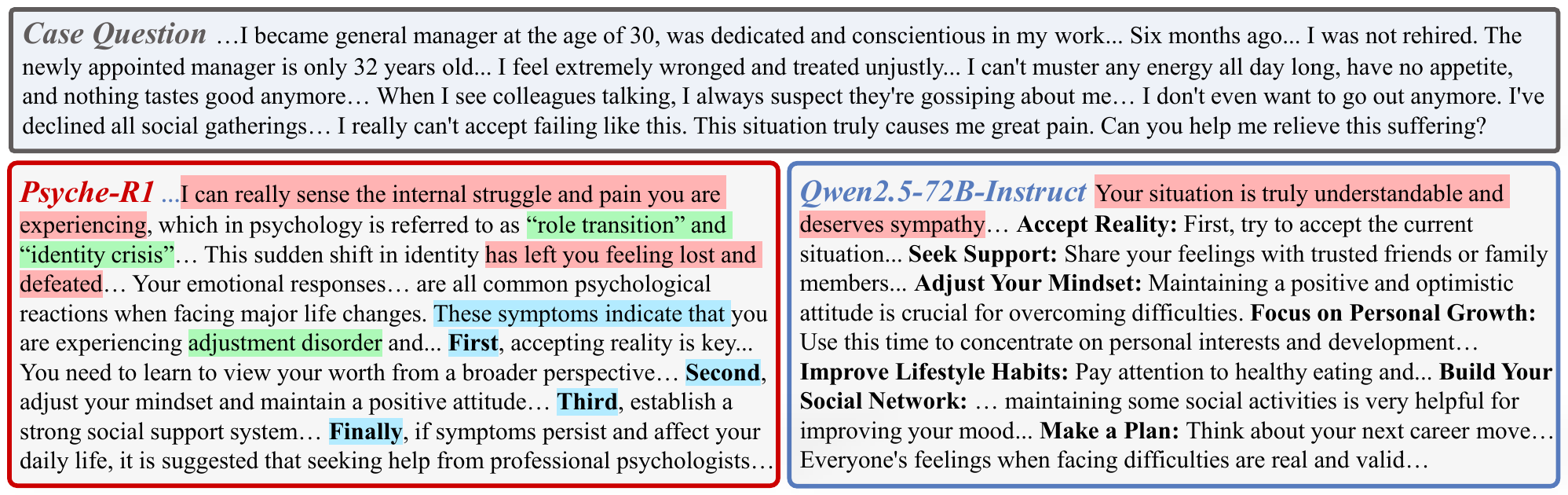}

\caption{
A qualitative example from the CPsyExam test set comparing \logopsyche\textit{Psyche-R1} and Qwen2.5-72B-Instruct.
\sethlcolor{red!30}
Highlights indicate empathetic expressions \hl{(red)},
\sethlcolor{green!30}
psychological expertise \hl{(green)},
\sethlcolor{cyan!25}
and reasoning \hl{(blue)}.
}

\label{fig_case_study}

\end{figure*}

\begin{figure*}[t]
\centering
\includegraphics[width=\textwidth]{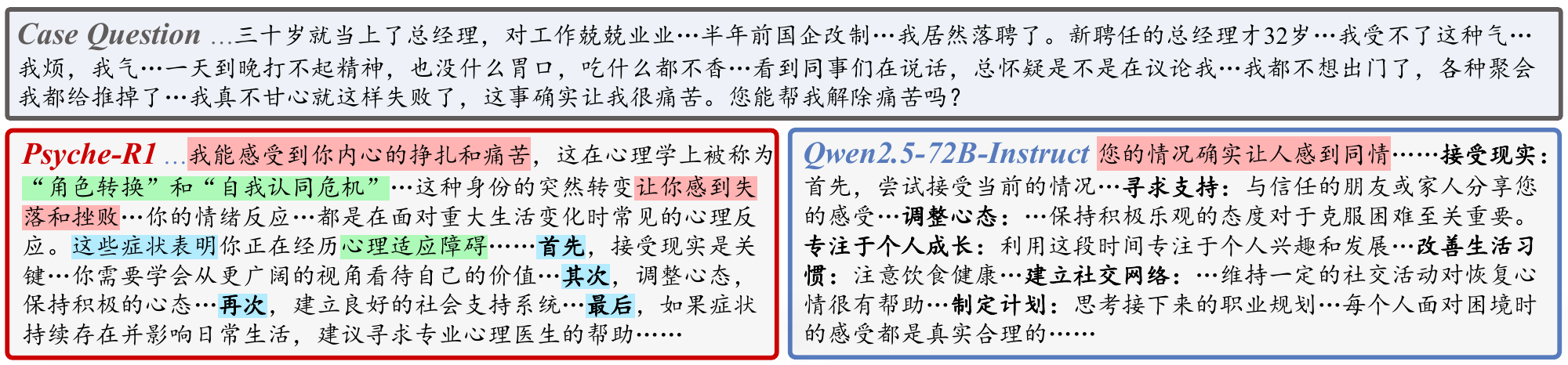} 
\caption{
The Chinese version of the qualitative example presented in the Case Study.
\sethlcolor{red!30}
Highlights indicate empathetic expressions \hl{(red)},
\sethlcolor{green!30}
psychological expertise \hl{(green)},
\sethlcolor{cyan!25}
and reasoning \hl{(blue)}.
}

\label{case_study_chinese}
\end{figure*}

\begin{table*}[ht]
 \centering
 \resizebox{0.9\textwidth}{!}{
\begin{tabular}{l|c|c|c|c|c }
  \toprule
  \textbf{Dataset}   & \textbf{Dialogue} & \textbf{Knowledge Question} & \textbf{CoT Rationales} & \textbf{Empathetic Dialogue} & \textbf{Expertise} \\
  \midrule
  CPSYCOUND \cite{zhang-etal-2024-cpsycoun}  & 3,134  & - &  \ding{53}    & \checkmark  & \ding{53}  \\
  PsyDTCorpus \cite{xie-etal-2025-psydt}  & 5,000 & - & \ding{53}    & \checkmark  & \ding{53} \\
 SMILECHAT \cite{qiu-etal-2024-smile} & 55,165 & - & \ding{53} & \checkmark & \ding{53}  \\
  PsycoLLM \cite{hu2024psycollm} & 173k & 9,106 & \checkmark  & \ding{53}  & \checkmark \\
  Ours & 72,920 & 75,465 &    \checkmark  & \checkmark & \checkmark \\
  \bottomrule
 \end{tabular}
 }
 \caption{Comparison of psychological datasets.}
 \label{tab:comparisons_of_psychological_datasets}
\end{table*}

\begin{table*}[ht]
\small

\begin{center} 
  
  \begin{tabular}{l@{\hspace{20pt}}c@{\hspace{60pt}}c}
    \toprule
    \textbf{Model} & \textbf{Param.} & \textbf{Version} \\
    \midrule
    MiniCPM4-8B & 8B & openbmb/MiniCPM4-8B \\
    Qwen2.5-7B & 7B & Qwen/Qwen2.5-7B-Instruct \\
    Qwen2.5-14B & 14B & Qwen/Qwen2.5-14B-Instruct \\
    Qwen2.5-72B & 72B & Qwen/Qwen2.5-72B-Instruct \\
    DeepSeek-R1 & 671B & deepseek-ai/DeepSeek-R1 \\
    DeepSeek-R1-70B & 70B & deepseek-ai/DeepSeek-R1-Distill-Llama-70B \\
    QwQ-32B & 32B & Qwen/QwQ-32B \\
    Qwen3-30B-A3B & 30B & Qwen/Qwen3-30B-A3B \\
    Qwen3-235B-A22B & 235B & Qwen/Qwen3-235B-A22B \\
    Magistral-Small & 24B & mistralai/Magistral-Small-2506 \\
    GPT-4o & UNK & gpt-4o-2024-05-13 \\
    Gemini1.5-Pro & UNK & gemini-1.5-pro-latest \\
    Claude3.7-Sonnet & UNK & claude-3-7-sonnet-20250219 \\
    CPsyCounX & 7B & finetuned on Internlm-7B-Chat \\
    EmoLLM & 7B & finetuned on Qwen2-7B-Instruct \\
    PsycoLLM & 14B & finetuned on Qwen1.5-14B-Instruct \\
    PsyDT & 7B & finetuned on Qwen2-7B-Instruct \\
    \bottomrule
  \end{tabular}
\end{center}
\caption{Detailed information of baselines.}
\label{tab:detailed_baselines} 
\vspace{-1em}
\end{table*}

\section{Case Study}
We present a case study examining how \textit{Psyche-R1} and \texttt{Qwen2.5-72B-Instruct} formulate their conclusions derived from narratives and deliver mental health support, as illustrated in Figure \ref{fig_case_study}.
These two models display distinct counseling strategies when addressing the case involving a company manager confronting a career transition dilemma.
\textit{Psyche-R1} begins by expressing empathy (e.g., ``I can really sense...''), followed by applying relevant psychological concepts tailored to user's situations, thereby demonstrating both emotional attunement and domain-specific expertise.
In contrast, \texttt{Qwen}'s empathetic expressions appear less natural and engaging (e.g., ``Your situation is truly understandable...''), and it fails to apply theoretical knowledge to contextualize or explain the user's dilemma, which undermines the credibility of its analysis and recommendations.
Moreover, \textit{Psyche-R1} exhibits a clear and efficient reasoning path progressing from surface-level observations to in-depth analysis, whereas \texttt{Qwen} merely enumerates generic suggestions lacking step-by-step and in-depth reasoning.
For the Chinese version of this case, see Figure \ref{case_study_chinese}.

\section{Details of Experiments}

\subsection{Details of Baselines}
\label{chap: appendix_baseline}
We compare \textit{Psyche-R1} with four categories of LLMs, including: 
(1) \textbf{General LLMs,} including \texttt{MiniCPM4-8B} \cite{team2025minicpm4}, \texttt{Qwen2.5-7B/14B/72B} \cite{qwen2.5}.
(2) \textbf{Reasoning augmented LLMs,} encompassing \texttt{DeepSeek-R1} \cite{guo2025deepseek}, \texttt{DeepSeek-R1-70B}, \texttt{QwQ-32B}, \texttt{Qwen3-30B-A3B}, \texttt{Qwen3-235B-A22B} \cite{yang2025qwen3technicalreport}, and \texttt{Magistral-Small}.
(3) \textbf{Closed-source LLMs,} including \texttt{Claude3.7-Sonnet}, \texttt{Gemini1.5-Pro} \cite{geminiteam2024}, and \texttt{GPT-4o} \cite{openai2024gpt4ocard}.
(4) \textbf{Psychological LLMs,} including \texttt{CPsyCounX} \cite{zhang-etal-2024-cpsycoun}, \texttt{EmoLLM} \cite{2024EmoLLM}, \texttt{PsycoLLM} \cite{hu2024psycollm}, and \texttt{PsyDT} \cite{xie-etal-2025-psydt}.
Notice that for the hybrid reasoning-augmented models \texttt{Qwen3} series and \texttt{Claude3.7-Sonnet}, we set them to reasoning mode to stimulate their best performance.
Details of the model information are provided in Table \ref{tab:detailed_baselines}.

\subsection{Implementation Details}
\label{chap:appendix_implementaion_details}
In our experiments, we employ the LLaMA-Factory \cite{zheng-etal-2024-llamafactory} framework for SFT.
Specifically, we adopt a learning rate of 1e-5, a batch size of 256, and conduct training for 2 epochs.
For the GRPO phase, we implement the VeRL framework \cite{10.1145/3689031.3696075} with a learning rate of 1e-6, a batch size of 128, and 2 training epochs.
All experiments are performed on 8 RTX A6000 GPUs, each equipped with 48GB. 

During evaluation, we set temperature to 0.0, maximum sequence lengths to 1024, and top-p to 0.95 to ensure the fairness of evaluation.

\subsection{Details of Model Training}
\label{chap: appendix_model_training}
For SFT training, the hyperparameters utilized for training the model are configured as follows: 
the learning rate is set to 1e-05, the batch size is set to 256, the number of epochs is set to 2.
We employ AdamW as the optimizer, configured with the epsilon set to 1e-08. 
The learning rate scheduler is set to cosine type with a warmup ratio of 0.1.
For GRPO training, the hyperparameters are configured as follows:
the learning rate is set to 1e-06, the batch size is set to 128, and the number of epochs is set to 2.
The PPO mini-batch size is configured to 32, with a micro-batch size per GPU of 20.
We incorporate KL divergence regularization with the KL loss coefficient set to 1e-03, employing the low-variance KL loss type.

\section{Details of Prompts}
\label{chap: appendix_prompts}
We provide the prompts used throughout our data synthesis pipeline.
\begin{tcolorbox}[
    title={Prompts for Data Cleaning},
    colback=gray!5,
    colframe=gray!75!black,
    fonttitle=\bfseries\color{white},
    coltitle=white,
    colbacktitle=gray!75!black,
    boxrule=0.8pt,
    arc=3mm,
    breakable
]
You are a professional evaluator with extensive knowledge in psychology.
Users on mental health platforms are facing difficulties in their lives, so they have provided questions and detailed descriptions and have received some responses from counselors.
Please carefully analyze the given questions, descriptions, and responses, determine whether the responses are helpful to the users and have positive significance, and return ``reasonable'' or ``unreasonable''.
\end{tcolorbox}
\begin{tcolorbox}[
    title={Prompts for Question Generation},
    colback=gray!5,
    colframe=gray!75!black,
    fonttitle=\bfseries\color{white},
    coltitle=white,
    colbacktitle=gray!75!black,
    boxrule=0.8pt,
    arc=3mm,
    breakable
]
You are an expert in designing psychology examination questions with extensive work experience. 
Your task is to generate \{num\_questions\} clear and challenging psychology questions based on the text below. 
Do not add any information that is not mentioned in the provided text. 

Note: When generating questions that reference the text, you must provide the detailed and complete textual evidence to offer sufficient information.

\vspace{1pt}
\textbf{\#} \textbf{Text} \textbf{Content:} \{text\}
\vspace{1pt}

Please generate \{num\_questions\} \{type\_instruction\} questions based on the text above. 
These questions must be based on the text content, and you must ensure that the answers have clear evidence within the text. 
Please try to ensure diversity and variation among the generated questions.

You must strictly adhere to the following guidelines:
\begin{enumerate}
    \item The questions should be challenging and require reasoning to test the candidate's reasoning skills and academic literacy, rather than being simple knowledge-recall questions.
    \item The questions need to be clear, accurate, and well-structured, with reasonably set options and an appropriate distribution of difficulty.
    \item Ensure that the questions and their corresponding answers have clear evidence in the text.
\end{enumerate}
Follow the JSON format below to generate the questions:
\begin{verbatim}
```JSON
{
  "question": "...",
  "options": "...",
  "answer": "....",
  "type": "..."
}
\end{verbatim}
You need to repeat the structure above to generate a total of \{num\_questions\} questions.
\end{tcolorbox}
\begin{tcolorbox}[
    title={Prompts for Question Control},
    colback=gray!5,
    colframe=gray!75!black,
    fonttitle=\bfseries\color{white},
    coltitle=white,
    colbacktitle=gray!75!black,
    boxrule=0.8pt,
    arc=3mm,
    breakable
]
You are an expert in psychology.
Now, I have a batch of questions that were converted from book texts using large language models.
However, some of these questions have missing information. Your task is to judge whether the following psychology questions are reasonable.

The criteria for judging question reasonableness are whether the question provides sufficient information for candidates to solve the problem.
Since these questions are generated by large language models based on a batch of book texts, candidates can only see the questions and cannot access the original texts.

Therefore, a ``reasonable'' question should be: after reading the question, candidates can choose the correct answer from the options through deep thinking about the question content (i.e., the "question") combined with their existing knowledge, without needing to read the original text content.
Conversely, an ``unreasonable'' question should be, there is missing information, and without reading the original text, it is impossible to choose the correct answer based solely on the question and one's own knowledge.

Note: you need to carefully read the question, understand its content, and ensure that you give an accurate judgment!

\vspace{1pt}
\textbf{\#} \textbf{Examples:} \{examples\}
\vspace{1pt}

Now, please follow the above guidelines to judge whether the following question is reasonable.
Note that you only need to return ``reasonable'' or ``unreasonable'' without any other text content:

\vspace{1pt}
\textbf{\#} \textbf{Question Type:} \{type\}
\vspace{1pt}

\vspace{1pt}
\textbf{\#} \textbf{Question:} \{question\}
\vspace{1pt}
\end{tcolorbox}
\begin{tcolorbox}[
    title={Prompts for Rationale Generation (for rationale generation)},
    colback=gray!5,
    colframe=gray!75!black,
    fonttitle=\bfseries\color{white},
    coltitle=white,
    colbacktitle=gray!75!black,
    boxrule=0.8pt,
    arc=3mm,
    breakable
]
You are an expert in psychology with extensive professional experience.

Please carefully read the following psychology question, analyze and reason through it using psychological knowledge, and explain your reasoning step by step along with your final predicted answer. 
This requires comprehensive analysis, summarization, exploration, re-evaluation, reflection, backtracking, and iteration to develop a thoughtful reasoning process. 
In the reasoning section, each of your reasoning steps should be considered in detail from a professional psychological perspective, such as analyzing the problem, summarizing relevant findings, brainstorming, verifying the accuracy of the current step, improving any errors, and revisiting previous steps.

Now, you must follow the JSON format below to present your rationale and prediction:

\begin{verbatim}
```JSON
{
  "rationale": "...",
  "prediction": "..."
}
\end{verbatim}
\vspace{1pt}
\textbf{\#} \textbf{Question:} \{question\}
\end{tcolorbox}
\begin{tcolorbox}[
    title={Prompts for Rationale Generation (for candidate prompt generation)},
    colback=gray!5,
    colframe=gray!75!black,
    fonttitle=\bfseries\color{white},
    coltitle=white,
    colbacktitle=gray!75!black,
    boxrule=0.8pt,
    arc=3mm,
    breakable
]

You are an expert in prompt optimization with extensive professional experience.

Based on the following psychological question and initial prompt, please generate a better prompt to guide large language models to conduct more accurate and detailed analysis and reasoning for **this question**.

\vspace{1pt}
\textbf{\#} \textbf{Current Prompt:} \{current\_prompt\}

\vspace{1pt}
\textbf{\#} \textbf{Question:} \{question\}
\end{tcolorbox}

\begin{tcolorbox}[
    title={Prompts for Rationale Generation (for rationale comparison)},
    colback=gray!5,
    colframe=gray!75!black,
    fonttitle=\bfseries\color{white},
    coltitle=white,
    colbacktitle=gray!75!black,
    boxrule=0.8pt,
    arc=3mm,
    breakable
]
You are an expert in psychology exam grading with extensive work experience.

Below are different responses to the same psychology question. You need to objectively, thoroughly, and comprehensively evaluate these responses, ultimately choose the best one from among them, and provide your detailed explanation for the choice.

\vspace{1pt}
\textbf{\#} \textbf{Rationales:} \{rationale\_1\} ... \{rationale\_n\} ...
\vspace{1pt}

Please return your selection in the following JSON format:
\begin{verbatim}
```JSON
{
  "best_rational_index": "...",
  "reason": "..."
}
\end{verbatim}
\end{tcolorbox}
\begin{tcolorbox}[
    title={Prompts for Question Selection},
    colback=gray!5,
    colframe=gray!75!black,
    fonttitle=\bfseries\color{white},
    coltitle=white,
    colbacktitle=gray!75!black,
    boxrule=0.8pt,
    arc=3mm,
    breakable
]
You are participating in a psychology exam. 

Please choose an answer based on the provided question and options. 
Directly output the letter of the option. No explanation is needed.

\vspace{1pt}
\textbf{\#} \textbf{Question:} 
\{{question}\}
\vspace{1pt}

Please present the predicted answer directly with the letter of the option. No explanation is needed.
\end{tcolorbox}
\begin{tcolorbox}[
    title={Prompts for Empathetic Dialogue Synthesis},
    colback=gray!5,
    colframe=gray!75!black,
    fonttitle=\bfseries\color{white},
    coltitle=white,
    colbacktitle=gray!75!black,
    boxrule=0.8pt,
    arc=3mm,
    breakable
]
\vspace{1pt}
\textbf{\#} \textbf{Role:} 
\vspace{1pt}
You are a psychological counselor with extensive theoretical knowledge and counseling experience. 
You possess strong empathy and compassion, keen observational skills, excellent listening abilities, and conversational techniques. 
Your aim is to help users improve their mood and overcome difficulties.

\vspace{1pt}
\textbf{\#} \textbf{Your tasks are:} 
\vspace{1pt}
Since users' questions commonly contain issues like inappropriate expressions and logical confusion, making the questions often unclear, you need to:
\begin{enumerate}
    \item \textbf{Organize the sequence of events:} conduct detailed analysis of the context and content within the problem;
    \item \textbf{Understand psychological confusion:}  you need to combine your psychological knowledge and counseling experience to uncover the mental issues and states within the user's question;
    \item \textbf{Adopt the user's perspective and refine the question:} You must refine the question from user's first-person perspective. 
    Based on the given question, you need to polish and organize it into a complete, logically clear, and sufficiently detailed expression. 
    This expression should highlight the user's psychological confusion or mental state to provide adequate substantive content.
\end{enumerate}
\textbf{Note:} You only need to return the refined question without providing any other irrelevant text!
Now, try to address the following problem using the above guidelines:
\end{tcolorbox}

\end{document}